\documentclass[journal]{IEEEtran}

\usepackage{cite}
\usepackage{amsmath,amssymb,amsfonts}
\usepackage{graphicx}
\graphicspath{{figs/}}
\DeclareGraphicsExtensions{.pdf,.png}
\usepackage{textcomp}
\usepackage{booktabs}
\usepackage{array}
\usepackage{multirow}
\usepackage{stfloats}
\usepackage{placeins}
\usepackage{balance}
\usepackage{url}
\usepackage[hidelinks]{hyperref}
\hypersetup{
  pdfauthor={Shuoqin Zhang, Tongtong Cheng, Xiru Gao, Jinzhuo Peng, Bin Zheng, Jiahao Tu, Ke Wang, Jia Pan, Zhe Hu, Kai Liu},
  pdftitle={EvoHIL: Self-Evolving Reward and Flow-Matched Policy Optimization for Robust Human-in-the-Loop Reinforcement Learning}
}

\newcommand{\EvoHIL}{EvoHIL}
\newcommand{\SER}{SER}
\newcommand{\AFS}{AFS}
\newcommand{\IEEEtablecaptionfont}{\fontencoding{T1}\fontfamily{ntxtlf}%
  \fontseries{m}\fontshape{sc}\selectfont}

\begin{document}
\bstctlcite{BSTcontrol}

\title{EvoHIL: Self-Evolving Reward and Flow-Matched Policy Optimization for Robust Human-in-the-Loop Reinforcement Learning}

\author{Shuoqin Zhang, Tongtong Cheng, Xiru Gao, Jinzhuo Peng,\\[-0.2ex]
Bin Zheng, Jiahao Tu, Ke Wang, Jia Pan, Zhe Hu\raisebox{0.6ex}{\scriptsize\textdagger}, and Kai Liu\raisebox{0.6ex}{\scriptsize\textdagger}%
\thanks{Shuoqin Zhang and Tongtong Cheng contributed equally to this work.}%
\thanks{Shuoqin Zhang, Xiru Gao, Bin Zheng, Jiahao Tu, Ke Wang, and Zhe Hu
are with the National Elite Institute of Engineering, Chongqing University,
Chongqing 401135, China.}%
\thanks{Tongtong Cheng and Kai Liu are with the College of Computer Science,
Chongqing University, Chongqing 400044, China.}%
\thanks{Jinzhuo Peng is with the College of Mechanical and Vehicle Engineering,
Chongqing University, Chongqing 400044, China.}%
\thanks{Jia Pan is with the Department of Computer Science, The University of
Hong Kong, Hong Kong, China.}%
\thanks{This work was conducted in collaboration with industry partner Chengdu
Anu Intelligence, Chengdu, China. Shuoqin Zhang and Zhe Hu are also affiliated
with Chengdu Anu Intelligence.}%
\thanks{\textdagger\ Corresponding authors: Zhe Hu and Kai Liu
(e-mail: jjhu1993@gmail.com; liukai0807@gmail.com).}%
\thanks{This work has been submitted to IEEE for possible publication. Copyright
may be transferred without notice, after which this version may no longer be
accessible.}}

\markboth{}{}

\maketitle

\begin{abstract}
Human-in-the-loop reinforcement learning (HIL-RL) enables robots to learn
contact-rich manipulation from limited real-world interaction, but deployment
exposes three coupled limitations: static visual reward models fail under scene
changes; independently sampled actions cause temporally inconsistent motion; and
vision-based policies remain sensitive to appearance shifts. We present \EvoHIL{},
a unified framework that adapts the reward model, action generator, and visual
domain within a staged human-in-the-loop learning process. First, self-evolving
reward (\SER{}) adapts the success classifier from human-confirmed positives and
provisional weak negatives. Second, Action Flow Stabilization (\AFS{}) generates
temporally coherent action chunks through flow matching, grounding policy updates
in executed action prefixes and demonstrated behavior. Third,
retention-aware offline fine-tuning replays relit interaction data while anchoring
the AFS actor--critic to prior behavior, adapting the visual domain without
additional robot interaction. Across six
manipulation tasks on Franka FR3 and SO-101 arms under a controlled lighting shift,
\EvoHIL{} improves task success, agreement with human-confirmation labels, motion
smoothness, and completion time relative to human-in-the-loop and imitation
baselines. Project page:
\href{https://anonymous4366.github.io/EvoHIL/}{https://anonymous4366.github.io/EvoHIL/}.
\end{abstract}


\section{Introduction}
\IEEEPARstart{H}{uman-in-the-loop} reinforcement learning provides a practical
approach for teaching precise, contact-rich manipulation directly on physical
robots. Recent systems combine online reinforcement learning with human
demonstrations and real-time teleoperated interventions to learn skills such as
connector insertion and surface manipulation from a modest amount of real-world
interaction~\cite{luo2025hilserl,luo2024serl,ball2023rlpd}. Human feedback is
particularly valuable when autonomous exploration is unsafe or inefficient: an
operator can demonstrate useful behavior, correct unsafe or ineffective actions,
and confirm task completion. However, these signals are commonly used to improve a
policy while the surrounding reward and action interfaces remain fixed. Continued
policy optimization therefore leaves limitations outside the policy update
unresolved.

Three limitations become important after deployment. A visual success classifier
is usually trained before online learning and then fixed as the terminal-reward
source. Later human confirmations reveal some of its errors but do not update it.
Appearance changes can consequently corrupt terminal labels and the Bellman targets
used for policy learning. The per-step Gaussian actor presents a separate problem:
it samples each command independently, although contact-rich manipulation often
requires coordinated corrections over several control steps. Sparse success rewards
provide little information about command variation between reward events. The
vision policy also remains tied to the appearance distribution observed during
interaction. An illumination change can alter its input without changing task
geometry or semantics, and further optimization on the original replay cannot
supply the missing visual coverage.

\begin{figure*}[!t]
\centering
\includegraphics[width=\textwidth]{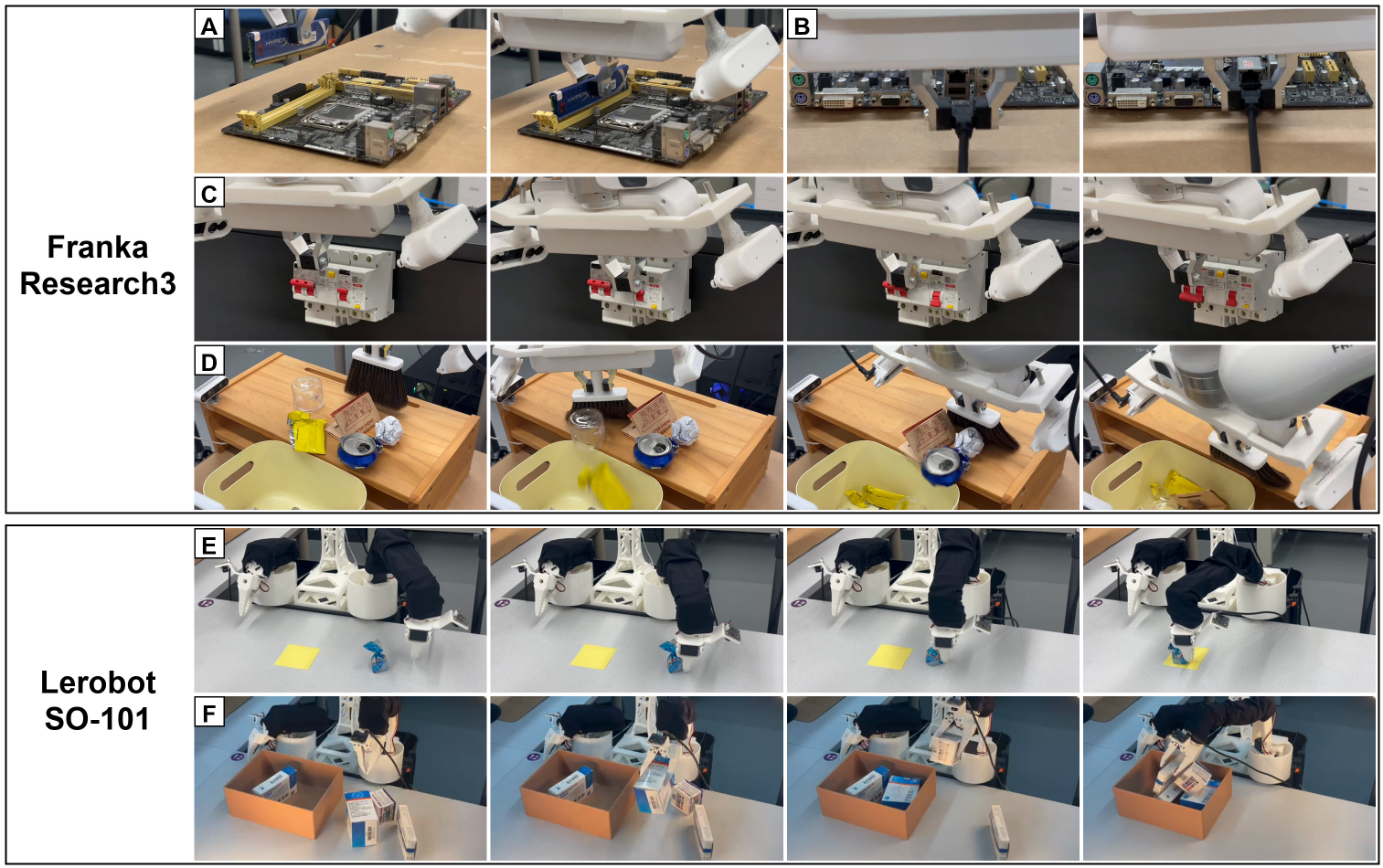}
\caption{Six real-robot manipulation tasks across two embodiments. A 7-DoF
Franka FR3 performs RAM insertion, USB insertion, table wiping, and circuit-breaker
operation; a low-cost SO-101 arm performs candy pushing and medicine-box stowing.}
\label{fig:scenes}
\end{figure*}

These limitations interact through the critic. An erroneous terminal label changes
the target used to evaluate behavior. An action-chunk actor must learn from the part of
a plan that was actually executed, and visual adaptation must retain the values and
behavior learned in the source domain. Joint fine-tuning without these distinctions
can create a positive self-labeling loop, apply a policy objective to unsuitable
replay data, or overwrite prior competence. Continued training is therefore not
enough; the reward, action, and visual interfaces require different supervision and
data paths. We study this problem on six tasks and two robot platforms
(Fig.~\ref{fig:scenes}) under predefined illumination changes.

\begin{figure*}[t]
\centering
\includegraphics[width=0.98\textwidth]{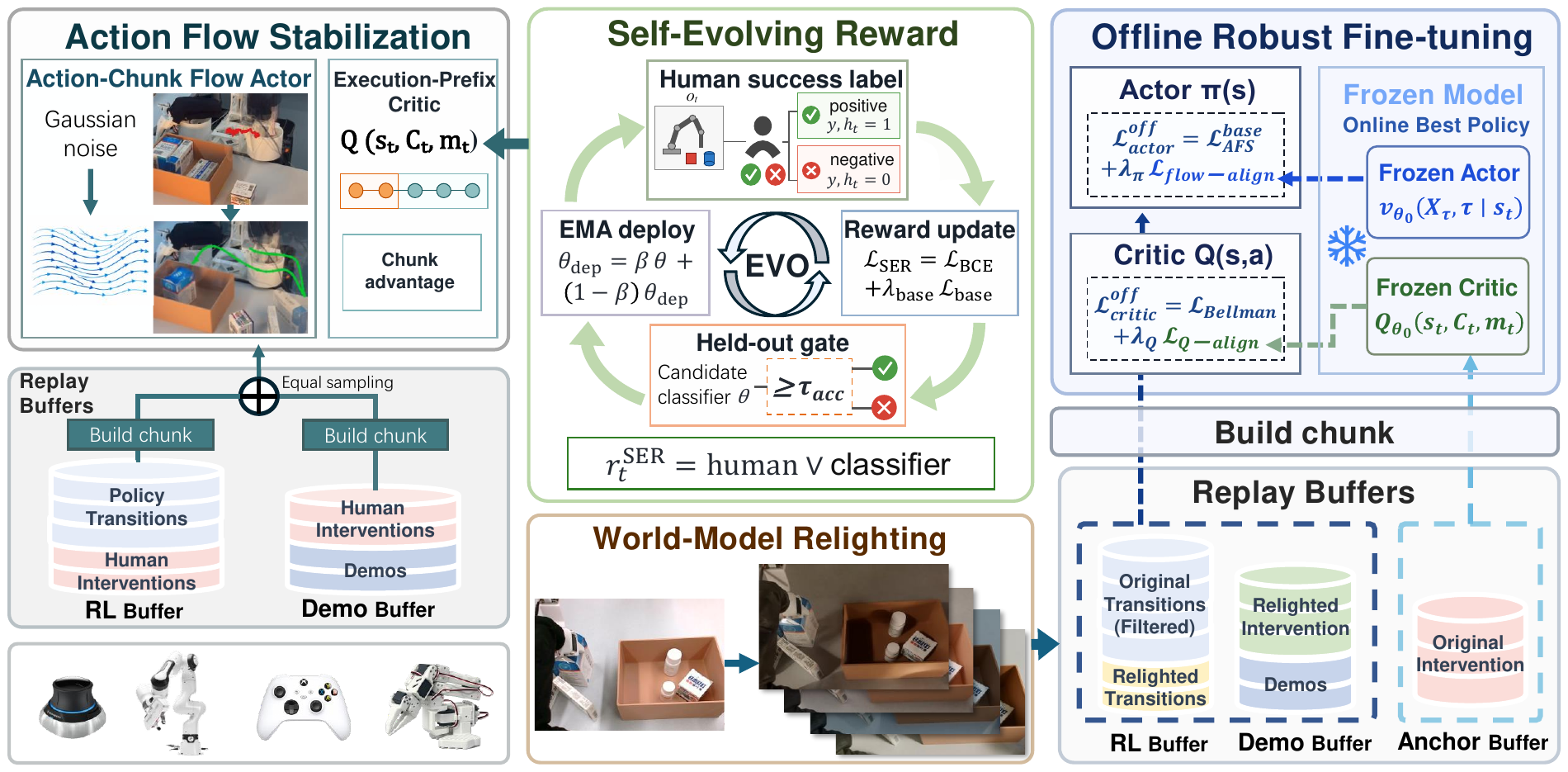}
\caption{Staged overview of \EvoHIL{}. During online interaction, \SER{} updates the
deployed success classifier from human confirmations, and \AFS{} learns action
chunks from source-tagged persistent replay.
Retention-aware offline fine-tuning subsequently combines source and externally
relit replay with frozen references for execution-prefix values and flow
velocities. The
three components share an actor--critic workflow but use distinct data paths and
training stages.}
\label{fig:overview}
\end{figure*}

We present \EvoHIL{}, a staged framework that assigns one adaptive mechanism to each
interface within a common actor--critic workflow
(Fig.~\ref{fig:overview}). Self-evolving reward (\SER{}) updates the deployed success
classifier from explicit confirmations and provisional weak negatives. Label-source
isolation prevents classifier predictions from becoming positive supervision, and
a gated EMA update limits abrupt changes to the deployed reward. Action Flow
Stabilization (\AFS{}) replaces independent commands with short flow-matched action
chunks. An execution-prefix critic evaluates the portion of each chunk that reaches
the robot, while source masks keep policy, demonstration, and intervention replay in
their designated objectives. A subsequent retention-aware offline phase combines
source observations with externally relit versions of recorded interaction. Frozen
references for execution-prefix values and flow velocities constrain adaptation
without additional robot data. The components are linked through the critic. \SER{}
changes subsequent reward targets, which alter the advantages used by \AFS{}. This
coupling does not introduce an auxiliary reward-shaping term.

This separation preserves the role of each supervision source: confirmations update
only the reward model, replay tags select the actor objective, and relit observations
enter only offline adaptation. The critic propagates revised success estimates under
retention constraints.

The contributions are summarized as follows.
\begin{itemize}
\setlength{\itemsep}{1pt}
\setlength{\parsep}{0pt}
\setlength{\topsep}{2pt}
\item \EvoHIL{} integrates adaptive reward supervision, action generation, and
visual-domain coverage through a common critic without changing the sparse reward
semantics.
\item \SER{} updates the reward model from confirmations already collected during
training. Confirmed frames are positive examples, whereas unconfirmed frames are
provisional weak negatives. Label-source isolation, a deployment gate, EMA updates,
and immutable replay rewards make the one-sided label bias explicit.
\item \AFS{} generates short action chunks with a flow-matching actor and evaluates
replay-based policy improvement on the same executed prefix as the critic. Expert
imitation and temporal regularization provide source-specific replay supervision.
\item Retention-aware offline fine-tuning adapts the AFS actor--critic to a new scene
using relit recordings without collecting more robot data. Anchors on
execution-prefix values and flow velocities limit deviations from source-domain
behavior.
\end{itemize}

Across all six tasks and both embodiments, the complete system is independently
evaluated with the same AFS actor family, execution-prefix critic, and
retention-aware offline phase. The online reward classifier recovers agreement
with the human-confirmation stream after each scene change, whereas the static
classifier remains poorly aligned. The flow actor reduces normalized command
irregularity, and the complete pipeline records the strongest selected-policy
success rates across the predefined illumination conditions. Component studies
separately examine reward adaptation, temporal regularity, and source-domain
retention; the resulting evidence supports robustness to the evaluated shift family
rather than unrestricted visual generalization.

\section{Related Work}
\label{sec:related}

\paragraph{Human-in-the-loop robot learning}
Vision-based real-robot RL requires sample-efficient and safe data
collection~\cite{kober2013rlrobotics,kalashnikov2018qtopt,ibarz2021trainrobot}.
Off-policy methods reuse interaction and pre-collected
data~\cite{fujimoto2018td3,kostrikov2022iql}; recent systems further combine replay
with demonstrations and
interventions~\cite{luo2024serl,luo2025hilserl,ball2023rlpd}, while DAgger variants
request
corrective expert actions~\cite{ross2011dagger,kelly2019hgdagger}. \EvoHIL{}
extends this setting by adapting the reward model and action representation while
preserving the source labels of replay segments.

\paragraph{Reward adaptation}
Robot rewards can be learned from preferences or visual
events~\cite{christiano2017preferences,fu2018vice}. PU reward learning uses unlabeled
states~\cite{xu2021purl}, and ARHI adapts a VLM-based dense reward from sparse
human guidance~\cite{zhou2026arhi}. \SER{} instead updates a binary terminal-success
classifier and accepts only explicit confirmations as trusted positives. Its
deployment gate addresses class imbalance and drift but does not remove
confirmation-selection
bias~\cite{guo2017calibration,he2009imbalanced,gama2014conceptdrift,ovadia2019datasetshift}.

\paragraph{Generative action policies}
Diffusion and implicit policies represent multimodal
actions~\cite{chi2023diffusionpolicy,florence2021ibc}, while flow matching learns a
continuous velocity field~\cite{lipman2023flowmatching,liu2023rectifiedflow}.
Action-chunk methods execute a predicted prefix before
replanning~\cite{zhao2023act,li2025actionchunking}. DPPO applies policy gradients to
diffusion policies~\cite{ren2025dppo}; ReinFlow introduces stochastic flow dynamics
for tractable online optimization~\cite{zhang2025reinflow}; FPO uses changes in the
conditional flow-matching loss~\cite{mcallister2026fpo}; and SAC Flow performs
off-policy optimization through velocity
reparameterization~\cite{zhang2026sacflow}. \AFS{} instead applies an FPO-style
surrogate only to
replayed policy segments and aligns it with an execution-prefix critic. This
surrogate is off-policy and is not an exact likelihood ratio.

\paragraph{Visual adaptation and temporal regularity}
Visual adaptation commonly uses augmentation or domain
randomization~\cite{laskin2020rad,kostrikov2021drq,tobin2017domainrand,hendrycks2020augmix}.
\EvoHIL{} relights recorded transitions with an external world
model~\cite{nvidia2025cosmostransfer}, retains source-domain
replay~\cite{rolnick2019clreplay}, and anchors execution-prefix values and flow
velocities at expert states. Separately, action regularization can reduce abrupt
commands~\cite{mysore2021caps}; \AFS{} combines finite differences with
cross-replan overlap consistency. The resulting metrics quantify normalized-command
regularity, not physical safety or human perception.

\section{Preliminaries and Problem Formulation}
\label{sec:prelim}

\subsection{Human-in-the-Loop Replay Learning}
We model vision-based manipulation as an MDP
$(\mathcal{S},\mathcal{A},P,r,\gamma)$, with state
$s_t=(I_t^{1:C},x_t)$, bounded end-effector action $a_t$, and objective
$\mathbb{E}[\sum_t\gamma^t r_t]$. The Gaussian reference backbone uses off-policy
SAC~\cite{haarnoja2018sac} and the single-step ensemble target
\begin{equation}
\label{eq:sac-target}
y_t=r_t+\gamma(1-d_t)\min_jQ_{\bar\theta,j}(s_{t+1},a_{t+1}),
\quad a_{t+1}\sim\pi_\theta(\cdot|s_{t+1}).
\end{equation}
Entropy regularization is applied to the Gaussian actor but not to this implemented
critic backup. Equation~\eqref{eq:sac-target} defines only the reference backbone;
all \EvoHIL{} results use the execution-prefix macro target of \AFS{}.

Following HIL-SERL and RLPD~\cite{luo2025hilserl,ball2023rlpd}, the learner samples
equally from online replay and a prior store containing demonstrations and online
human interventions. Both support off-policy critic learning; the prior store is
therefore not an offline dataset in the strict sense. In \AFS{}, source tags further
restrict the FPO-style surrogate to policy segments and BC-flow to expert segments.

\subsection{Deployment Drift and Fixed Interfaces}
\label{sec:two-sources}
Let $q_{\mathrm{tr}}$ and $q_{\mathrm{dep}}$ denote training and deployment
observation distributions. Appearance changes can produce
$q_{\mathrm{dep}}\ne q_{\mathrm{tr}}$, corrupting the output of a fixed visual
success model and its Bellman targets. A single-step diagonal Gaussian also lacks a
joint distribution over future actions and an explicit temporal constraint.
Continued policy optimization alone neither updates the reward classifier nor
expands visual-domain coverage. We therefore adapt the reward and action interfaces
under $q_{\mathrm{dep}}$ and use recorded relit replay to limit degradation under
$q_{\mathrm{tr}}$ without additional robot interaction.

\section{Method}
\label{sec:method}
\EvoHIL{} retains the human-in-the-loop actor--critic loop described in
Section~\ref{sec:prelim}, but allows the reward model and action representation to
change while retaining the source labels of replay segments. A retention-aware
substrate limits source-domain degradation during policy adaptation. \SER{} updates
the reward model, and \AFS{} replaces the Gaussian actor with a flow-matched action
policy. The two components interact through the critic rather than through reward
shaping.

\subsection{Overview: A Self-Evolving Closed Loop}
\label{sec:method-overview}
Fig.~\ref{fig:overview} illustrates the learning loop. The actor interacts with the
environment using the deployed reward model and stores transitions in the replay
buffer. Human success confirmations and interventions are recorded as they occur.
On the learner side, the reward model is updated using explicit human confirmations
and conservative weak-negative labels (\SER{}) and is then broadcast to the actor.
\AFS{}
uses persistent replay and prior data for its off-policy execution-prefix critic,
expert imitation, smoothness, and FPO-style updates. The FPO-style update is
restricted to policy-generated segments and their executed action prefixes. The
reward model
does not provide an auxiliary shaping term to the actor.
Instead, it determines the reward stored in each transition and affects the actor
only through the critic. A retention-aware substrate is applied in a subsequent
offline phase to limit drift of the critic and actor from behavior learned on the
training distribution. The three components form one workflow but are not updated
simultaneously.

\subsection{Retention-Aware Visual Adaptation}
\label{sec:method-substrate}
Continued policy adaptation must preserve previously acquired competence. The
substrate supports this objective at two levels: the data used to train the critic
and explicit anchors on the execution-prefix critic and flow actor.

\paragraph{Transition-preserving relit replay}
To provide critic coverage under shifted visual conditions without collecting
additional robot interaction, we reuse recorded trajectories and relight only
their images. For a transition $\tau_t=(o_t,a_t,r_t,o_{t+1},d_t)$ with
$o_t=(I_t^{1:C},x_t)$, an external relighter produces images
$\tilde I_{t}^{1:C}$ under a new lighting condition. We then construct
\begin{equation}
\label{eq:relit}
\tilde\tau_t=\big((\tilde I_t^{1:C},x_t),\,a_t,\,r_t,\,(\tilde I_{t+1}^{1:C},x_{t+1}),\,d_t\big),
\end{equation}
keeping the action, reward, and termination unchanged,
$\tilde a_t=a_t,\ \tilde r_t=r_t,\ \tilde d_t=d_t$. Because only the visual channel
is altered, $\tilde\tau_t$ can be reused as an off-policy transition under the
geometry-preservation assumption below: the relit data introduce a visual-domain
shift without changing the recorded control labels.
This construction assumes that the external relighter preserves task geometry,
object state, and the temporal correspondence between $I_t$ and $I_{t+1}$. The
relighter is external to \EvoHIL{} and may not reproduce every extreme illumination
effect. Rendering errors limit the visual coverage of the relit replay. Another
relighter can be substituted without changing \eqref{eq:relit}, the replay buffers,
or the retention objectives.

\paragraph{Dual-domain replay with a retention ratio}
Let $\mathcal{D}^{\mathrm{src}}_{\mathrm{on}\setminus\mathrm{intv}}$ denote
source-light online replay after removing transitions that match the source-light
intervention set, and let $\mathcal{D}^{\mathrm{relit}}_{\mathrm{on}}$ denote the
relit version of the full online replay. The replay buffer mixes the two as
\begin{equation}
\label{eq:retention-mix}
\begin{split}
p_{\mathcal{B}_{\mathrm{replay}}}(\tau)
={}&\alpha\,p\!\left(\tau\,|\,\mathcal{D}^{\mathrm{src}}_{\mathrm{on}\setminus\mathrm{intv}}\right)\\
&+(1-\alpha)\,p\!\left(\tau\,|\,\mathcal{D}^{\mathrm{relit}}_{\mathrm{on}}\right),
\qquad \alpha=0.75 .
\end{split}
\end{equation}
The source term retains Bellman coverage of the training distribution with weight
$\alpha$, while the relit term covers the shifted distribution with weight
$1-\alpha$.
The prior-data buffer
$\mathcal{B}_{\mathrm{prior}}=\mathcal{D}^{\mathrm{src}}_{\mathrm{demo}}\cup
\mathcal{D}^{\mathrm{relit}}_{\mathrm{intv}}$ collects expert behavior only:
original-light demonstrations and relit human interventions. A reference mask
$z_i=\mathbb{1}[\tau_i\in\mathcal{B}_{\mathrm{prior}}]$ marks which samples carry
expert behavior. Only these expert samples are used as anchors. Failure and
recovery states remain governed by the Bellman updates of
$\mathcal{B}_{\mathrm{replay}}$.

\paragraph{Anchored regularization}
On the masked expert samples, we regularize the current networks toward the
reference parameters $\theta_0$, which are frozen at the start of fine-tuning. The
complete system uses
the same \AFS{} chunk representation for the current and frozen networks. Let
$\mathbf C_i$ denote the padded executed prefix, $\mathbf m_i$ its validity mask,
$\mathbf U_i$ the recorded action chunk, and $\mathbf b_i$ its chunk-validity mask.
The critic anchor distills the execution-prefix $Q$-values of the frozen critic.
Defining
$Q_{\theta,j}^i=Q_{\theta,j}(o_i,\mathbf C_i,\mathbf m_i)$,
\begin{equation}
\label{eq:critic-anchor}
\mathcal{L}_{\mathrm{c\text{-}anc}}
= \lambda_Q\,\frac{\sum_i z_i\,\mathrm{mean}_j
\big(Q_{\theta,j}^i-Q_{\theta_0,j}^i\big)^2}
{\sum_i z_i+\epsilon}.
\end{equation}
For samples from a separate reference stream, the corresponding bootstrap value
also uses the frozen critic $Q_{\theta_0}$, evaluated on the same next prefix sampled
from the current actor. Non-reference rows retain the Polyak target in
Eq.~\eqref{eq:afs-target}.
The actor anchor aligns the current and frozen velocity fields for the same expert
chunk, flow time $\tau_i$, noise $\epsilon_i$, and interpolation point
$\mathbf X_i=(1-\tau_i)\epsilon_i+\tau_i\mathbf U_i$. With
$v_{\theta,i,k}=v_\theta(X_{i,k},\tau_i|o_i)$,
\begin{equation}
\label{eq:actor-anchor}
\mathcal{L}_{\mathrm{a\text{-}anc}}
=\lambda_\pi
\frac{\sum_{i,k}z_ib_{i,k}
\big\|v_{\theta,i,k}-\mathrm{sg}\,v_{\theta_0,i,k}\big\|_2^2}
{d_a\sum_{i,k}z_ib_{i,k}+\epsilon},
\end{equation}
where $d_a$ is the action dimension and padded tokens are excluded by $b_{i,k}$.
The total critic and actor objectives add these terms to their respective losses.
Retention-aware fine-tuning is offline in the strict interaction sense: it uses
recorded source and relit transitions and collects no new rollouts from the current
behavior policy, human labels, or expert actions. It is a required training phase
of \EvoHIL{} rather than optional post-processing. Unless an ablation explicitly
removes it, every result labeled \EvoHIL{} includes online SER--AFS learning followed
by this phase. Its execution-prefix critic is trained off-policy, while the actor
uses expert BC-flow, smoothness, and the actor anchor; the replay-based FPO term is
disabled. This phase preserves the chunk values and action-generation behavior of
the frozen references without introducing a Gaussian action-mode or distributional
anchor that is not implemented by the AFS actor. We set $\lambda_Q=0.2$ and
$\lambda_\pi=0.1$.

\subsection{Self-Evolving Reward (\SER{})}
\label{sec:method-ser}
A fixed visual success model can become inaccurate under deployment drift
(Section~\ref{sec:two-sources}). \SER{} updates a binary terminal-success
classifier online from positive-only human confirmation feedback. Its objective is
a conservative detector aligned with the confirmation stream, not an unbiased
estimate of latent task success. The update follows a strict labeling rule that
prevents self-reinforcing errors in positive labels.

\paragraph{Deployed reward}
The deployed reward is computed using a classifier with success probability
$p_{\mathrm{dep}}(o)=\sigma(z_{\theta_{\mathrm{dep}}}(o))$. To suppress isolated
false positives, the classifier contributes to the reward only when its output
exceeds a high threshold $\tau$ for $K_{\mathrm{conf}}$ consecutive frames. Let
$h_t^+\in\{0,1\}$ denote an explicit human success confirmation at transition $t$,
$c_t$ the number of consecutive threshold exceedances, and
$g_t=\mathbb{1}[c_t\ge K_{\mathrm{conf}}]$ the resulting indicator:
\begin{equation}
\label{eq:ser-count}
c_t=\begin{cases}
0, & h_t^+=1,\\
c_{t-1}+1, & h_t^+=0,\ p_{\mathrm{dep}}(o_{t+1})>\tau,\\
0, & \text{otherwise.}
\end{cases}
\end{equation}
The reward and termination are
\begin{equation}
\label{eq:ser-reward}
r_t=\mathbb{1}[\,h_t^+=1 \ \lor\ g_t=1\,],
\qquad d_t=d_t^{\mathrm{env}}\ \lor\ r_t.
\end{equation}
When \SER{} is enabled, the environment wrapper supplies only the human event and
does not run a second static classifier; hence only the explicit human event or the
deployed \SER{} classifier can emit the terminal success reward.

\paragraph{Label-source isolation and one-sided label noise}
Let $s_{t+1}^\star\in\{0,1\}$ denote the latent task-success state of the next
observation. Every observed next frame is assigned the proxy target
\begin{equation}
\label{eq:ser-label}
\tilde y_t = h_t^+.
\end{equation}
Even when a classifier prediction triggers a reward, the training label remains
$0$ in the absence of an explicit confirmation. Classifier
predictions therefore cannot independently become positive training labels. This
isolation prevents a confident false positive from labeling its input as successful,
thereby blocking a direct positive self-training loop.

We express the proxy relation in positive--unlabeled
terms~\cite{xu2021purl}. Let $Y(o)=s^\star\in\{0,1\}$ denote latent success,
$C(o)=h^+\in\{0,1\}$ denote confirmation, and
$q(o)=\Pr(C{=}1\mid Y{=}1,o)$ denote the confirmation-selection probability.
Assuming that confirmations are valid positives,
\begin{equation}
\label{eq:ser-selection}
\Pr(C{=}1\mid o)=q(o)\Pr(Y{=}1\mid o).
\end{equation}
If $q(o)$ is constant, the proxy preserves the ranking of latent success; if it
depends on the scene, task phase, or operator state, even this ranking need not be
preserved. An absent or delayed confirmation therefore does not prove failure.
Most $h_t^+{=}0$ samples are nonterminal observations, but this class can also
contain delayed, omitted, or classifier-first successes. We treat $h_t^+{=}1$ as a
trusted positive and $h_t^+{=}0$ only as a provisional weak negative. This
asymmetric choice controls false positives that can terminate an episode and
corrupt later TD targets, at the cost of downward and potentially state-dependent
bias. Accordingly, \SER{} is evaluated by agreement with the confirmation stream,
not by unobserved task-success accuracy.

\paragraph{Conservative online update}
Each update augments trusted positives with random crops and photometric jitter,
samples provisional negatives at a fixed negative-to-positive ratio, and minimizes
a weighted cross-entropy loss together with a consistency term relative to the
baseline classifier $\theta_0$, which is frozen at the start:
\begin{equation}
\label{eq:ser-loss}
\mathcal{L}_{\SER}(\theta)=\mathcal{L}_{\mathrm{BCE}}(\theta)
+\lambda_{\mathrm{base}}\,\mathbb{E}\big\|z_\theta(o)-\mathrm{sg}\,z_{\theta_0}(o)\big\|^2 .
\end{equation}
A new classifier is accepted only if its balanced accuracy exceeds a threshold on
a held-out split containing both proxy classes. This gate measures generalization
of the conservative confirmation proxy; it is not a claim of unbiased latent
success accuracy. If either the training or validation split lacks a proxy class,
the update is not deployed. The deployed parameters are updated by
an exponential moving average,
$\theta_{\mathrm{dep}}\leftarrow(1-\beta)\theta_{\mathrm{dep}}+\beta\theta$, which
limits the rate of change in the reward distribution observed by the critic.
Default values are $\tau=0.95$, $K_{\mathrm{conf}}=2$, negative-to-positive ratio $5$,
$\lambda_{\mathrm{base}}=0.1$, held-out fraction $0.25$, balanced-accuracy
acceptance threshold $0.92$, and $\beta=0.1$. Positive augmentation and balanced
validation reduce class-frequency effects, while baseline consistency, the
deployment gate, and EMA limit drift; these safeguards do not make
\eqref{eq:ser-label} an unbiased success label.
Rewards already stored in replay are not rewritten, which provides the critic with
a stable reward history.
This update avoids a circular label path. Human events define the labels, and the
held-out gate and EMA control which candidate becomes the deployed classifier. The
deployed classifier affects only rewards collected after deployment; neither its
predictions nor those rewards rewrite the training labels or historical replay.

\subsection{Action Flow Stabilization (\AFS{})}
\label{sec:method-afs}
\AFS{} replaces the single-step Gaussian policy with a conditional flow-matching
policy that predicts a short action chunk by integrating a learned velocity field.
Its policy update uses the FPO-style surrogate~\cite{mcallister2026fpo} discussed in
Section~\ref{sec:related}. An off-policy critic evaluates executed action prefixes,
and the FPO-style update is restricted to prefixes generated by the policy. Expert
BC-flow uses only demonstration and intervention segments, while the smoothness
term is aligned with the executed sequence. These updates draw from persistent
replay. Only the retention phase is offline in the sense that it collects no new
robot interaction. By predicting coordinated chunks rather than independent commands,
\AFS{} reduces temporal irregularity in the executed trajectory.

\paragraph{Flow policy and sampling}
A velocity field $v_\theta(\mathbf X_\tau,\tau\,|\,s)$ is conditioned on the encoded
observation $s$, flow time $\tau\in[0,1]$, and intermediate action chunk
$\mathbf X_\tau$. For a recorded chunk
$\mathbf U_i=(a_{i,0},\ldots,a_{i,H-1})$ with validity mask $b_{i,k}$, we draw
$\epsilon_i^n\sim\mathcal{N}(0,I)$ and $\tau_i^n\sim\mathcal{U}(0,1)$, form
$X_{i,k}^n=(1-\tau_i^n)\epsilon_{i,k}^n+\tau_i^na_{i,k}$, and use the masked,
horizon-normalized conditional-flow loss. With
$u_{i,k}^n=a_{i,k}-\epsilon_{i,k}^n$ and
$\ell_{i,k}^n=\|v_\theta(X_{i,k}^n,\tau_i^n|s_i)-u_{i,k}^n\|_2^2$,
\begin{equation}
\label{eq:cfm}
\hat{\mathcal{L}}_{\mathrm{cfm},i}(\theta;\mathbf b_i)
=\frac{1}{N}\sum_{n=1}^{N}
\frac{\sum_k b_{i,k}\eta^k\ell_{i,k}^n}
{d_a\sum_k b_{i,k}\eta^k+\epsilon},
\end{equation}
where $\eta=0.95$ discounts distant chunk positions and $d_a$ is the action
dimension. This decay emphasizes near-term actions, which are more likely to be
executed before receding-horizon replanning. The flow policy produces an action
chunk of length $H$. At inference, we integrate the velocity field from a Gaussian
latent using $K_{\mathrm{flow}}$ Euler steps, clip the result to the action bounds,
and execute the first $E<H$ actions before replanning. This
receding-horizon scheme is consistent with recent evidence that predicting and
executing short action chunks can improve temporal coherence and exploration in
RL~\cite{li2025actionchunking,zhao2023act}. Executing $E$ actions rather than the
entire chunk preserves policy responsiveness. The smoothness term described below
uses the horizon overlap between successive chunks. We use short chunks
($H{=}8$, $E{=}2$ for the FR3 insertion and breaker tasks, and up to $H{=}24$,
$E{=}4$ for the longer SO-101 tasks; Table~\ref{tab:task-configs}). For the stow
task, the learned gripper command is
represented as one dimension of the same flow-generated chunk; the environment
thresholds this dimension during execution.

\begin{figure}[t]
\centering
\includegraphics[width=\columnwidth]{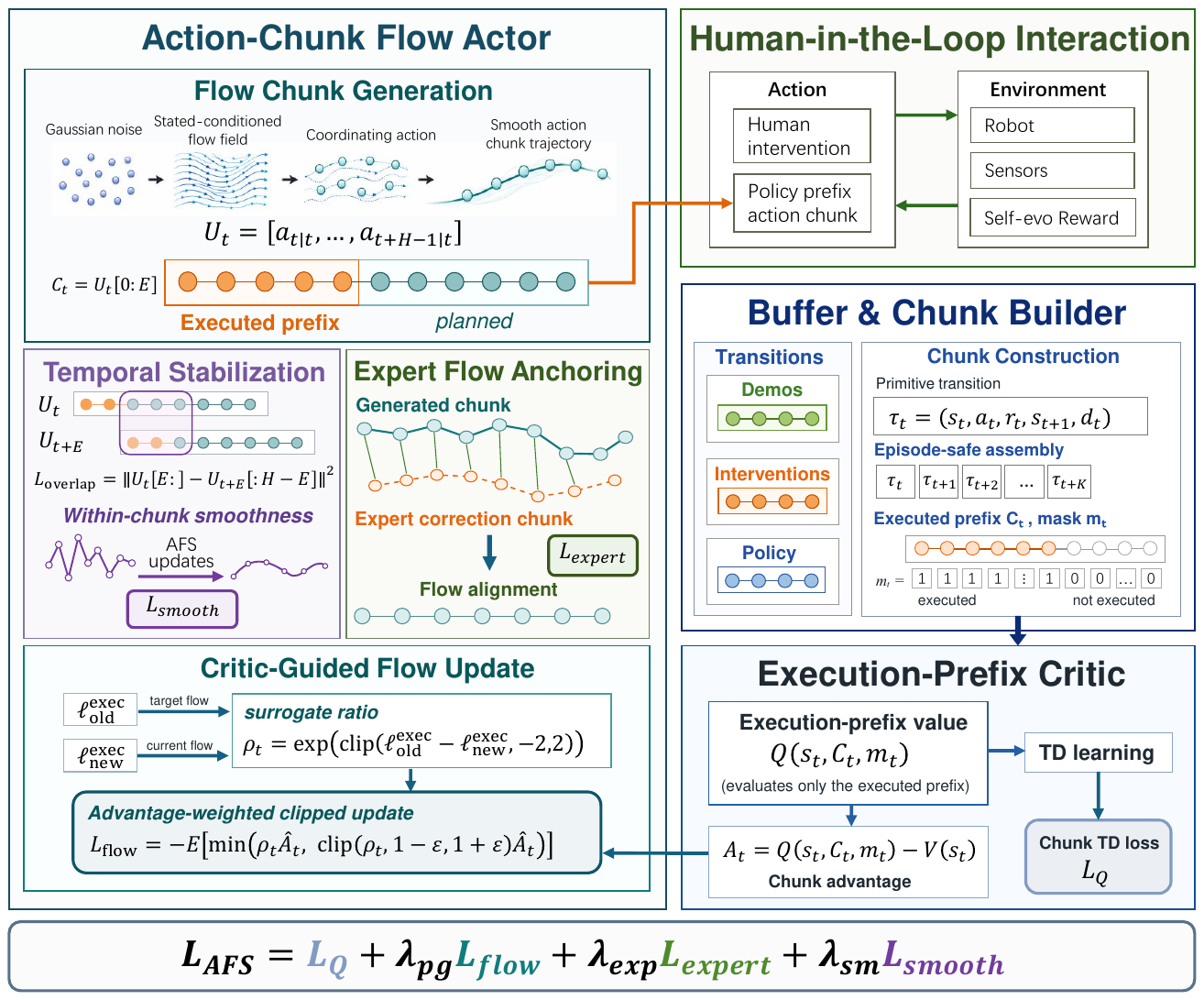}
\caption{Detailed architecture of \AFS{}. The flow actor predicts an $H$-step
action chunk and executes the first $E<H$ actions before replanning. Policy,
demonstration, and intervention transitions are assembled into chunks that respect
episode boundaries for the execution-prefix critic. Policy-only segments support
the executed-prefix FPO-style update; the execution-prefix macro TD critic, expert
BC-flow objective, and temporal-smoothness objective use their designated replay
subsets.}
\label{fig:flowschematic}
\end{figure}

\paragraph{Source-aware replay segments}
Each primitive transition stores its trajectory index, chunk-start flag, policy
version, and action source. At sampling time, consecutive actions form an
$H$-step recorded segment $\mathbf U_i$ without crossing an episode, source, or
policy-version boundary. A policy segment is eligible for FPO only when its first
$E$ actions are available or when it terminates earlier. Demonstration,
intervention, and random warm-up segments are excluded from FPO. Because eligible
policy segments remain in persistent replay after collection, the resulting
surrogate is off-policy with respect to the current actor. The version tag prevents
policy versions from being mixed within a segment but does not make the surrogate
on-policy.

\paragraph{Off-policy critic and advantage}
\AFS{} constructs an execution-aligned macro transition from primitive replay. Let
$e_i\le E$ denote the number of executable actions before an episode boundary,
$\mathbf C_i=(a_{i,0},\ldots,a_{i,e_i-1})$ its zero-padded prefix, and
$\mathbf m_i$ its validity mask. The sampler aggregates each primitive reward once,
\begin{equation}
R_i^{(e_i)}=\sum_{k=0}^{e_i-1}\gamma^k r_{i,k},
\end{equation}
and trains the execution-prefix ensemble critic with the variable-step target
\begin{equation}
\label{eq:afs-target}
y_i^{(e_i)}=R_i^{(e_i)}+\gamma^{e_i}(1-d_i^{(e_i)})
\min_j Q_{\bar\theta,j}(s_{i+e_i},\mathbf C'_{i+e_i},\mathbf 1),
\end{equation}
where $\mathbf C'_{i+e_i}$ is the first $E$ actions of a newly sampled flow chunk.
The next prefix is sampled from the current actor, and $Q_{\bar\theta,j}$ denotes
the Polyak-averaged target critic.
The critic evaluates $Q_j(s_i,\mathbf C_i,\mathbf m_i)$ rather than a one-step
action. Because these macro transitions are sampled repeatedly from persistent
policy, demonstration, intervention, and relit replay, the critic update is
off-policy. A macro transition never crosses an episode, data-source, or policy-
version boundary. A nonterminal prefix shortened by such a boundary retains its
bootstrap term, whereas an incomplete nonterminal policy prefix is excluded from
FPO.

For each sampled segment, the actor advantage uses
$Q_{\mathrm{data},i}=\min_jQ_j(s_i,\mathbf C_i,\mathbf m_i)$ and
$V_\theta(s_i)=M^{-1}\sum_m\min_jQ_j(s_i,\pi_{\theta}^{(m)}(s_i)_{0:E},\mathbf 1)$.
The difference $Q_{\mathrm{data},i}-V_\theta(s_i)$ is standardized over the mixed
minibatch and clipped to $[-5,5]$ to obtain $A_i$. The policy mask selects the
entries used by FPO, whereas the expert mask selects the entries used to weight
BC-flow.

\paragraph{Flow policy-gradient surrogate}
Each replay entry retains an $H$-step recorded action segment, but FPO is evaluated
only on its executed prefix. Specifically, let
$p_{i,k}=b_{i,k}\mathbb{I}[k<E]$ denote the executed-prefix mask. Before computing
actor gradients, the learner fixes the flow times and noise for the batch and
evaluates a reference loss with the Polyak-averaged target actor,
$\ell_i^{\mathrm{ref}}=
\hat{\mathcal{L}}_{\mathrm{cfm},i}(\bar\theta;\mathbf p_i)$. The current loss is
$\ell_i^{\mathrm{new}}=
\hat{\mathcal{L}}_{\mathrm{cfm},i}(\theta;\mathbf p_i)$ under the same flow times
and noise. The surrogate ratio is
\begin{equation}
\rho_i(\theta)=
\exp\!\big(\mathrm{clip}(\ell_i^{\mathrm{ref}}-\ell_i^{\mathrm{new}},-2,2)\big).
\end{equation}
Let $\mathcal P$ denote the eligible policy rows in the sampled minibatch, and
define the clipped ratio as $\bar\rho_i(\theta)=\mathrm{clip}
(\rho_i(\theta),1-\varepsilon,1+\varepsilon)$. The policy surrogate is
\begin{equation}
\label{eq:fpo}
\begin{aligned}
\mathcal{L}_{\mathrm{FPO}}^{\mathrm{pol}}(\theta)
&=-\frac{1}{|\mathcal P|}\sum_{i\in\mathcal P}
\min\!\left\{\rho_i(\theta)A_i,\bar\rho_i(\theta)A_i\right\}.
\end{aligned}
\end{equation}
The ratio is a CFM-loss surrogate rather than an exact normalized flow-likelihood
ratio. Fixing flow times and noise reduces within-batch comparison variance but does
not provide a valid importance-sampling ratio, an explicit KL constraint, or a
monotonic-improvement guarantee. The target actor is a slowly moving optimization
reference, not the behavior
policy that generated a replayed action. FPO acts on the same executed prefix
evaluated by the critic; the unexecuted suffix receives no FPO gradient and is
trained only through expert imitation and temporal regularization when applicable.

\paragraph{Expert imitation and replay smoothness}
All valid demonstration and intervention chunks contribute to flow imitation, with
greater weights assigned to chunks with positive advantages. With
$w_i^{\mathrm{exp}}=\mathrm{is\_exp}_i
\exp(\mathrm{clip}(\max(A_i,0),0,3))$, the full-chunk loss is
\begin{equation}
\label{eq:bcflow}
\mathcal{L}_{\mathrm{BC}}^{\mathrm{exp}}(\theta)=
\frac{\sum_i w_i^{\mathrm{exp}}\,
\hat{\mathcal{L}}_{\mathrm{cfm},i}(\theta;\mathbf b_i)}
{\max(\sum_i w_i^{\mathrm{exp}},1)} .
\end{equation}
Here, $\mathrm{is\_exp}_i$ is one for demonstration and intervention chunks and
zero otherwise, and $\mathbf b_i$ is the full-chunk validity mask.
A smoothness term acts on the predicted chunk and combines three penalties: the
first difference $\|a_{k+1}-a_k\|^2$ and second difference
$\|a_{k+2}-2a_{k+1}+a_k\|^2$ within a chunk, and an overlap-consistency penalty that
matches the shared portion of the chunks predicted at states $t$ and $t{+}E$ under
common latent noise. The first two suppress jitter and curvature inside the executed
segment. The overlap term penalizes disagreement between successive replans over
their shared actions, thereby maintaining continuity of the combined trajectory
across chunk boundaries. These states come from persistent replay; no old-policy
loss is constructed for this auxiliary term. The online actor objective is
\begin{equation}
\label{eq:afs-actor}
\begin{aligned}
\mathcal{L}_{\mathrm{actor}}^{\mathrm{online}}(\theta)
={}&\lambda_{\mathrm{FPO}}(t)\mathcal{L}_{\mathrm{FPO}}^{\mathrm{pol}}(\theta)
+\lambda_{\mathrm{BC}}(t)\mathcal{L}_{\mathrm{BC}}^{\mathrm{exp}}(\theta)\\
&+\lambda_{\mathrm{sm}}\mathcal{L}_{\mathrm{smooth}}^{\mathrm{rep}}(\theta).
\end{aligned}
\end{equation}
During retention-aware offline fine-tuning, the FPO term is disabled and the actor
anchor in \eqref{eq:actor-anchor} is added.
Training begins with a $5$k-update critic/BC warm-up, followed by a $5$k linear FPO
ramp. During this period, the BC multiplier decays from $1$ to $0.5$. The default
values are $K_{\mathrm{flow}}=5$, $N=4$,
$\varepsilon=0.1$, $M=4$, base imitation weight $0.2$, and smoothness weight $0.05$.

\subsection{Coupling Through the Critic, Not Reward Shaping}
\label{sec:method-couple}
\SER{} and \AFS{} interact only through the critic and are not connected by an
additional shaping term. Updates to the deployed reward model can alter subsequently
stored rewards and, consequently, the macro TD target in \eqref{eq:afs-target} and
the advantage and flow-actor updates:
\begin{equation}
\label{eq:chain}
\text{proxy reward}\ \rightarrow\ \text{TD target}\ \rightarrow\ \text{advantage}\ \rightarrow\ \text{actor}.
\end{equation}
This design preserves the task objective. Reward shaping must be potential-based to
leave the optimal policy unchanged~\cite{ng1999rewardshaping}; an unverified
shaping term can alter the task. Improving the reward source instead of
adding a shaping term retains the same terminal-success semantics. \SER{} and
\AFS{} can therefore operate independently or jointly. When both are enabled,
\AFS{} uses rewards collected after the deployment of \SER{}.

\section{Experiments}
\label{sec:experiments}
The experiments examine whether the retention-aware substrate balances adaptation
to shifted scenes with retention in the source domain
(Section~\ref{sec:exp-substrate}), and whether the reward classifier recovers its
agreement with the human-confirmation proxy after a scene change
(Section~\ref{sec:exp-ser}). We then compare the temporal regularity and task
outcomes of the flow and Gaussian actors (Section~\ref{sec:exp-afs}) and evaluate
the adapted systems across predefined illumination changes on
both embodiments (Section~\ref{sec:exp-crossembodiment}). A controlled lighting
change serves as the representative environmental shift. Table~\ref{tab:60shift}
reports means over $60$ independently reset evaluation trials for every method and
scene. Other evaluation counts are stated with the corresponding result; these
trials measure rollout-level variability and are not independent training runs. The
available
evaluation summaries report mean recorded episode duration as completion time; the
run-level censoring records needed for a time-to-success analysis are unavailable,
so time is treated as a descriptive secondary metric and interpreted jointly with
SR. An episode containing any human intervention is scored as unsuccessful, and
intervention rate is the fraction of intervention-controlled primitive steps among
all executed primitive steps. The robot is reset after every episode. Reported
checkpoints were selected under the archived evaluation protocol, which does not
document a separate set for checkpoint selection; the comparisons are therefore
descriptive and may include checkpoint-selection effects. The abbreviations ``og'' and ``re''
denote evaluations under original and shifted lighting, respectively.

\subsection{Experimental Setup}
\label{sec:exp-setup}
\begin{figure}[t]
\centering
\includegraphics[width=\columnwidth]{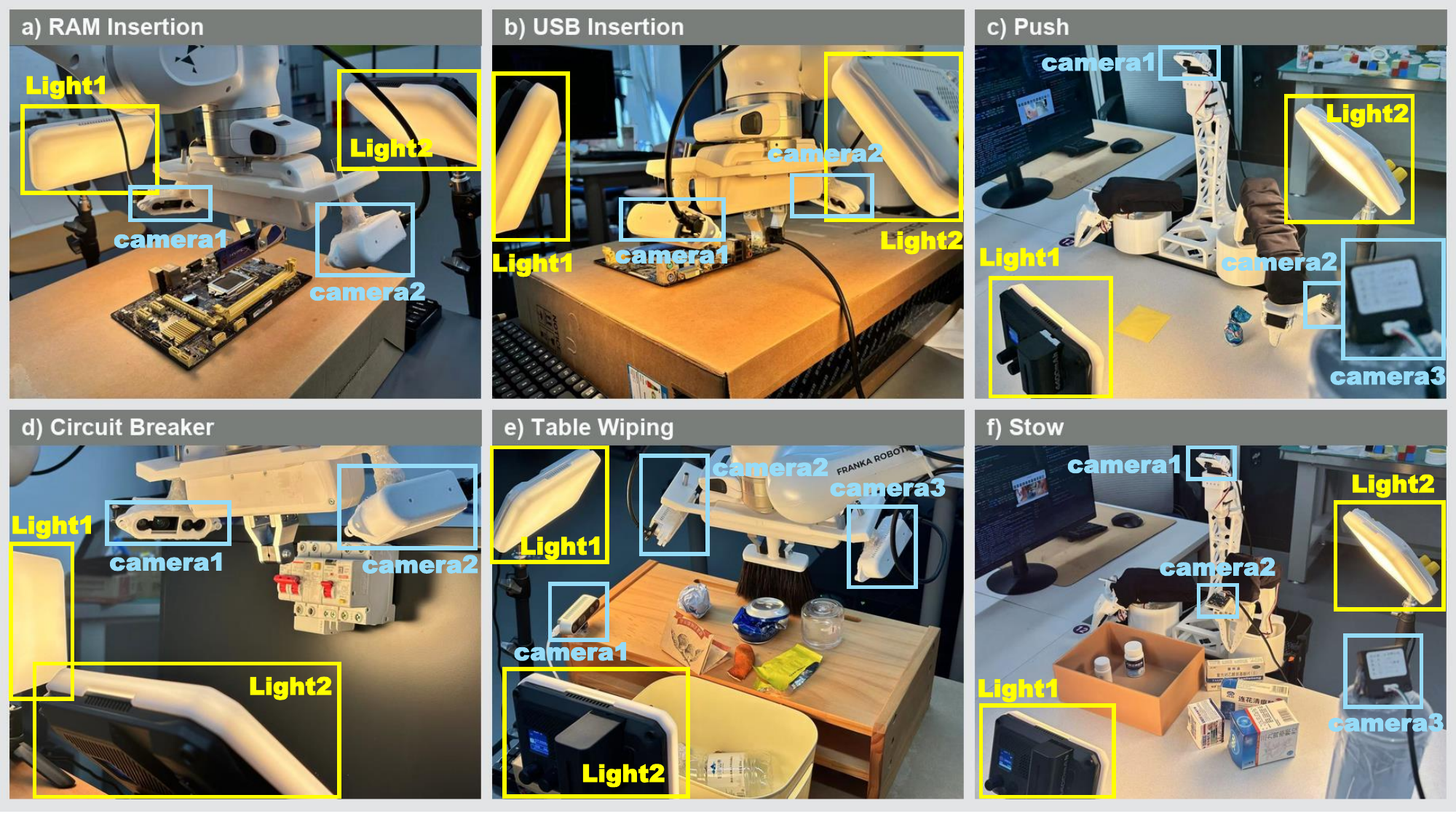}
\caption{Camera placements for the six scenes. The FR3 tasks use wrist-mounted
cameras (a side camera is added for table wiping); the SO-101 tasks use top and
side cameras, with an added wrist camera for stowing.}
\label{fig:cameras}
\end{figure}

We evaluate \EvoHIL{} on six manipulation tasks across two embodiments. A 7-DoF
Franka FR3
performs RAM insertion, USB insertion, table wiping, and circuit-breaker operation,
each with a 6-D end-effector action and a fixed gripper. A low-cost SO-101 arm
performs candy pushing with a fixed gripper and medicine-box stowing, for which
\AFS{} jointly generates the arm command and the thresholded gripper dimension.
Fig.~\ref{fig:cameras}
shows the camera placements. We vary a composite illumination-shift index from
$0\%$ to $100\%$ and report the $60\%$ operating point as the primary cross-domain
condition. The index orders ten predefined combinations of brightness, color
temperature, shadow, and reflection; it is not a calibrated percentage change in
lux. Because the recorded protocol does not include a held-out illumination set, the
sweep evaluates targeted adaptation and stress conditions rather than zero-shot
generalization to unseen illumination. We compare
against HIL-SERL~\cite{luo2025hilserl}, HG-DAgger~\cite{kelly2019hgdagger}, behavior
cloning (BC), IBRL~\cite{hu2024ibrl}, and ACT~\cite{zhao2023act} under matched
numbers of demonstrations, real-robot interaction budgets, sensor inputs, and a
common evaluation protocol. Every entry in Fig.~\ref{fig:lighting-gradient} and
Table~\ref{tab:60shift} was independently reevaluated for this study. Each entry
labeled \EvoHIL{} uses the same SER--AFS policy family and includes retention-aware
offline fine-tuning.
Table~\ref{tab:task-configs} provides the task-specific setups and hyperparameters.

The control interface runs at $10\,\mathrm{Hz}$, and cropped RGB observations are
resized to $128\!\times\!128$ pixels before encoding. All robot evaluations are
operator-supervised. Commands are clipped to task-specific Cartesian workspace or
joint-step bounds, and the operator can override the policy. The experiments measure
robot
task outcomes and do not involve human behavioral outcomes. Device-specific
inference latency was not retained as a separate evaluation record.

The offline phase is part of the method and adds no environment steps, human
confirmations, or expert actions. The comparison therefore matches demonstrations,
real-robot interaction, and sensing, but not relit observations, offline update
counts, or compute: these resources are used only by \EvoHIL{}. The baselines retain
their published objectives, so the main comparison is an end-to-end system
comparison rather than a resource-matched causal estimate of the contribution of
each component. Section~\ref{sec:exp-substrate} instead reports the online Gaussian
reference and ablates replay mixing and both anchors; Appendix
Table~\ref{tab:relight-cost} reports relighting cost.

\subsection{Shift Adaptation and Source-Domain Retention}
\label{sec:exp-substrate}
These studies use USB insertion on the FR3, where both an original-light and a
shifted-light evaluation are available for every configuration. To isolate the
retention substrate from the new action representation, these independent studies
use the single-step Gaussian SAC backbone, a frozen feature anchor, and a frozen
Gaussian action-mean anchor. The complete system retains the same replay interface
but replaces these backbone-specific anchors with the AFS-compatible objectives in
Eqs.~\eqref{eq:critic-anchor} and~\eqref{eq:actor-anchor}. These results assess the
replay and retention design rather than the AFS-specific anchor formulation.

\paragraph{Complementary Effects of the Two Anchors}
The feature and Gaussian action-mean anchors address complementary failure modes.
Table~\ref{tab:loss-ablation} shows that neither anchor is sufficient in isolation.
The feature-anchor variant records shifted-light success of
$0.97$, but original-light success decreases to $0.57$ and completion time
increases to $7.69\,\mathrm{s}$. Feature retention alone does not provide the best
balance in this run. Conversely, the action-mean-anchor variant records
original-light success of $1.00$ with a completion time of $3.05\,\mathrm{s}$, but
success in the relit domain remains $0.80$, with a completion time of
$6.16\,\mathrm{s}$. With both terms, success reaches $1.00$ in both domains, and
the joint variant achieves the shortest completion times in the table
($2.75\,\mathrm{s}$ og and
$2.48\,\mathrm{s}$ re). Within this USB study, the joint configuration provides
the strongest observed balance between source retention and shift adaptation.

\begin{table}[!t]
\centering
\caption{\IEEEtablecaptionfont Anchor Ablation on SAC Retention for USB Insertion (FR3); OG/RE:
Original/Shifted Light; One Selected Policy from One Training Run}
\label{tab:loss-ablation}
\footnotesize
\setlength{\tabcolsep}{3.8pt}
\begin{tabular}{@{}lcccc@{}}
\toprule
Variant & og SR $\uparrow$ & og T $\downarrow$ & re SR $\uparrow$ & re T $\downarrow$ \\
\midrule
No anchor (standard SAC)        & 0.77 & 5.30 & 0.67 & 3.95 \\
Feature anchor only             & 0.57 & 7.69 & 0.97 & 3.45 \\
Action-mean anchor only         & \textbf{\ensuremath{\mathbf{1.00}}} & 3.05 & 0.80 & 6.16 \\
Both anchors                    & \textbf{\ensuremath{\mathbf{1.00}}} & \textbf{\ensuremath{\mathbf{2.75}}} & \textbf{\ensuremath{\mathbf{1.00}}} & \textbf{\ensuremath{\mathbf{2.48}}} \\
\bottomrule
\end{tabular}
\end{table}

\begin{figure}[t]
\centering
\includegraphics[width=\columnwidth]{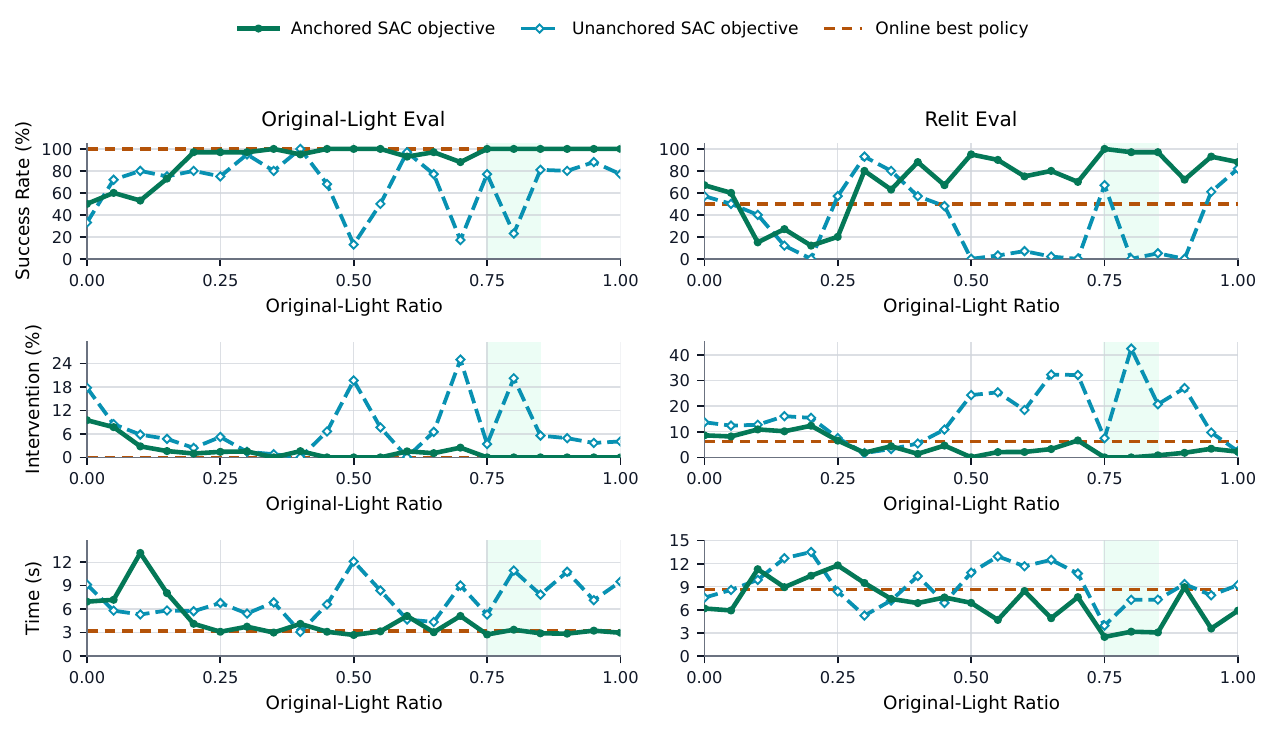}
\caption{Sweep of the original-domain retention ratio $\alpha$ for SAC fine-tuning
on USB insertion (FR3), with the best online Gaussian policy as a dashed
reference, across success rate, intervention rate, and completion time. Both
extremes perform poorly; the anchored objective is stable over a broad interior
band and has its best observed result at $\alpha{=}0.75$ in this run.}
\label{fig:buffer-ratio}
\end{figure}

\paragraph{Selection of the Retention Ratio}
Fig.~\ref{fig:buffer-ratio} presents a sweep of the original-domain retention ratio
$\alpha$. The observed curves indicate that anchoring reduces sensitivity to this
ratio. Without anchoring, the SAC fine-tuning objective is highly sensitive to
$\alpha$.
Shifted-light success is $0.00$--$0.07$ across much of the range
($\alpha{=}0.2,0.7,0.8,0.9$), and original-light success varies between $0.03$ and
$0.97$. With anchoring, the complete objective maintains original-light success at
or near $1.00$ across the entire
$\alpha{\in}[0.35,1.0]$ band and keeps shifted-light success in the $0.72$--$1.00$
range, peaking at $\alpha{=}0.75$ with $1.00/1.00$ success and an intervention rate
of zero. Both extremes remain suboptimal. At $\alpha{=}0$, the critic lacks Bellman
coverage of the training domain ($0.50/0.67$ with anchoring). At $\alpha{=}1.0$,
the buffer contains no relit data, and shifted-light success remains $0.88$. The
best observed operating point is therefore in the interior of the tested range, and the anchored
objective has a wider high-performing region in this sweep. We use $\alpha=0.75$
for the reported \EvoHIL{} experiments.

\begin{figure}[t]
\centering
\includegraphics[width=\columnwidth]{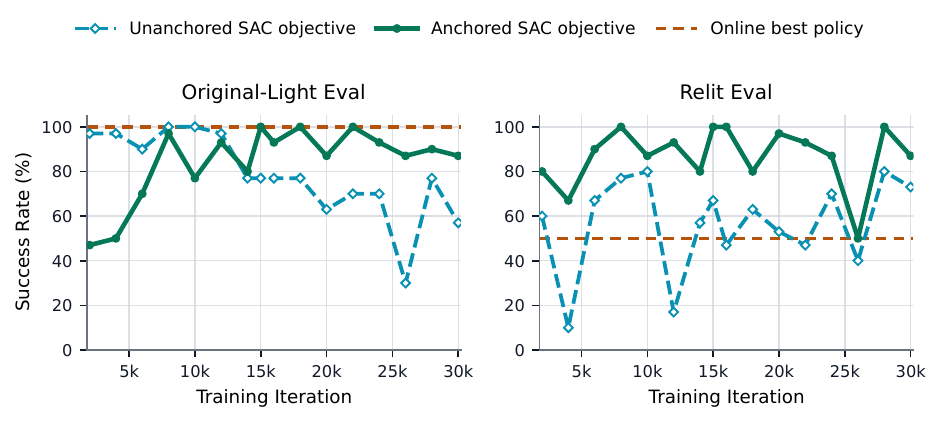}
\caption{Success rate during fine-tuning on USB insertion (FR3) for the anchored
and unanchored SAC objectives under original and shifted lighting, with the best
online policy as a dashed reference. The anchored objective remains stable and
exceeds the online-policy reference of $0.50$ on the shifted domain, whereas the
unanchored objective oscillates.}
\label{fig:loss-iter}
\end{figure}

\paragraph{Comparison with the Online Gaussian Reference}
Fig.~\ref{fig:loss-iter} tracks success over $30$k offline fine-tuning iterations.
The unanchored SAC objective is unstable on the shifted domain, oscillating between
$0.10$ and $0.80$ from one checkpoint to the next and decaying on the original
domain late
in training (to $0.30$ near $26$k), indicating source-domain degradation in the
unanchored run. By contrast, the anchored objective remains at or above
$0.77$ on the original domain after roughly $8$k iterations, exceeds $0.9$ at most
reported checkpoints, and reaches $1.00$ on the shifted domain at several
checkpoints ($8$k, $15$k, $16$k, $28$k). The best online policy obtained without
this substrate reaches $1.00$ under the original lighting but only $0.50$ under
shifted lighting. At the selected checkpoint, offline fine-tuning with relit replay
reaches $1.00$ in both domains without additional robot interaction. This
single-run comparison shows retention at the selected checkpoint, not a general
guarantee against forgetting.

\paragraph{Replay and Anchors Are Complementary}
The $2{\times}2$ study in Table~\ref{tab:joint-ablation} separates the observed
effects of the two mechanisms. The buffer mixture without anchors
reaches only $0.77/0.67$ (og/re), and the anchors without the mixture
($\alpha{=}0$) reach $0.50/0.67$. Both results are below the
$1.00/1.00$ achieved by the combination, and the corresponding completion times
are two to three times longer. In this ablation, the mixture supplies cross-domain
coverage, whereas the anchors are associated with better source retention during
adaptation.

\begin{table}[!t]
\centering
\caption{\IEEEtablecaptionfont Joint Replay--Anchor Ablation for USB Insertion (FR3); OG/RE:
Original/Shifted Light; One Selected Policy from One Training Run}
\label{tab:joint-ablation}
\footnotesize
\setlength{\tabcolsep}{3.8pt}
\begin{tabular}{@{}lccccc@{}}
\toprule
Replay & Anchors & og SR $\uparrow$ & og T $\downarrow$ & re SR $\uparrow$ & re T $\downarrow$ \\
\midrule
$\alpha{=}0$    & --         & 0.33 & 9.13 & 0.57 & 7.58 \\
$\alpha{=}0$    & \checkmark & 0.50 & 6.95 & 0.67 & 6.18 \\
$\alpha{=}0.75$ & --         & 0.77 & 5.30 & 0.67 & 3.95 \\
$\alpha{=}0.75$ & \checkmark & \textbf{\ensuremath{\mathbf{1.00}}} & \textbf{\ensuremath{\mathbf{2.75}}} & \textbf{\ensuremath{\mathbf{1.00}}} & \textbf{\ensuremath{\mathbf{2.48}}} \\
\bottomrule
\end{tabular}
\end{table}

\subsection{\SER{}: Confirmation-Label Agreement Recovers Across Scene Shifts}
\label{sec:exp-ser}
These studies use candy push and medicine-box stow on the SO-101, where the
deployed scene changes several times during training.

\begin{figure}[!t]
\centering
\includegraphics[width=\columnwidth,trim={197bp 5bp 198bp 9bp},clip]{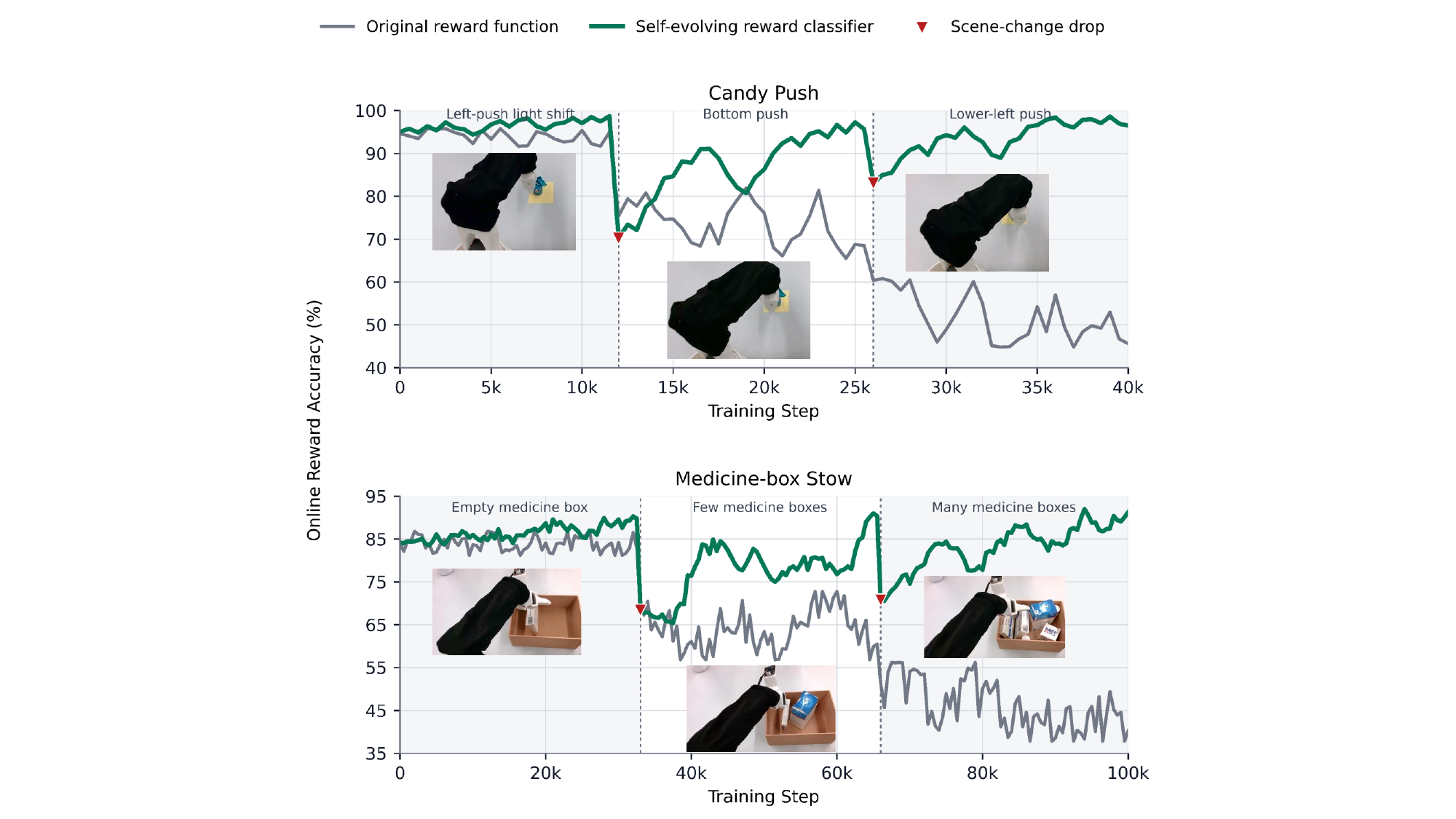}
\caption{Agreement with human-confirmation proxy labels for the deployed \SER{}
classifier and the original static reward function during training on candy push
($40$k steps, three scenes) and medicine-box stow ($100$k steps, three scenes).
Dotted markers indicate scene changes. Each point uses a random balanced sample of
at most $256$ frames from the cumulative live proxy-label buffers. Adjacent points
can therefore share frames, and unconfirmed true successes may be counted as
errors; these single-run curves do not estimate unbiased task-success accuracy.}
\label{fig:ser-acc}
\vspace{-18pt}
\end{figure}

\paragraph{Online Recovery After Each Scene Transition}
Fig.~\ref{fig:ser-acc} compares agreement with the human-confirmation proxy for the
self-evolving classifier and the original static reward function throughout
training. At each logging point, the metric is balanced accuracy computed on a
random balanced sample of at most $256$ frames from the accumulated
live positive and provisional-negative buffers. The sample is neither temporally
held out nor labeled independently for task success. At each scene change, the
agreement of the self-evolving classifier initially decreases and then recovers as confirmed
positives and provisional negatives accumulate. By contrast, the agreement of the
original reward function decreases and remains low. For candy push, with scene
transitions near $12$k and $26$k steps, the proxy agreement of the self-evolving
classifier begins near $95\%$, recovers after both transitions, and reaches
approximately $96\%$ at the end of training, with a peak near $99\%$. The
corresponding agreement of the original reward function decreases to approximately
$46\%$. For the longer-horizon medicine-box stow task, with transitions near $33$k
and $66$k steps during a $100$k-step run, the proxy-label agreement of the
self-evolving classifier decreases to
approximately $65\%$ after a transition and recovers to approximately $91\%$ by the
end of training. The agreement of the original reward function decreases to
approximately $40\%$. The self-evolving classifier therefore recovers agreement in
each new scene using human-confirmed positives and unconfirmed weak negatives,
whereas the static classifier remains poorly aligned with that stream after a
shift. Recovery is slower and more variable for the longer and more cluttered stow
task than for the push task, indicating that the recovery rate depends on scene
difficulty and the frequency of human confirmations.

\paragraph{Conservative Updates Without Self-Relabeling}
The recovery of proxy-label agreement after each scene shift is consistent with the
label-source isolation in \eqref{eq:ser-label}: classifier
outputs cannot relabel their inputs as positive and inflate agreement through a
direct self-training path. The metric nevertheless uses the same conservative label
definition as training. It therefore measures agreement with the confirmation
stream, not latent task-success accuracy, and the deployment gate cannot correct the
state-dependent selection bias formalized in \eqref{eq:ser-selection}. The
task-level results provide a separate closed-loop measure, but they do not establish
calibration against independent success labels.

\subsection{\AFS{}: Temporal Regularity and Task Outcomes}
\label{sec:exp-afs}
These studies use candy push and medicine-box stow on the SO-101 and compare the
\AFS{} flow actor with the Gaussian actor of the base human-in-the-loop system. The
flow actor applies the FPO-style surrogate to executed prefixes of policy-generated
replay segments. Its execution-prefix critic, expert BC-flow objective, and
smoothness objective use their designated persistent-replay subsets, as described in
Section~\ref{sec:method-afs}.

We compute the smoothness metrics over a 64-step window of executed normalized
actions
$\{a_t\}_{t=1}^{T}$. The first-difference metric is
$D_1=(T-1)^{-1}\sum_{t=2}^{T}\|a_t-a_{t-1}\|_2$, and the second-difference metric
is $D_2=(T-2)^{-1}\sum_{t=3}^{T}\|a_t-2a_{t-1}+a_{t-2}\|_2$. The high-frequency
power ratio $P_{\mathrm{HF}}$ is the action-signal power above half the Nyquist
frequency divided by total non-DC power. It measures rapid command variation;
smaller values indicate less high-frequency activity. We define the aggregate
smoothness score as
\begin{equation}
\label{eq:smooth-score}
S_{\mathrm{smooth}}=\frac{1}{1+D_1+D_2+P_{\mathrm{HF}}}.
\end{equation}
Lower values of $D_1$, $D_2$, and $P_{\mathrm{HF}}$, together with a higher value
of $S_{\mathrm{smooth}}$, indicate greater temporal regularity of the normalized
commands.

\begin{figure}[!t]
\centering
\includegraphics[width=\columnwidth]{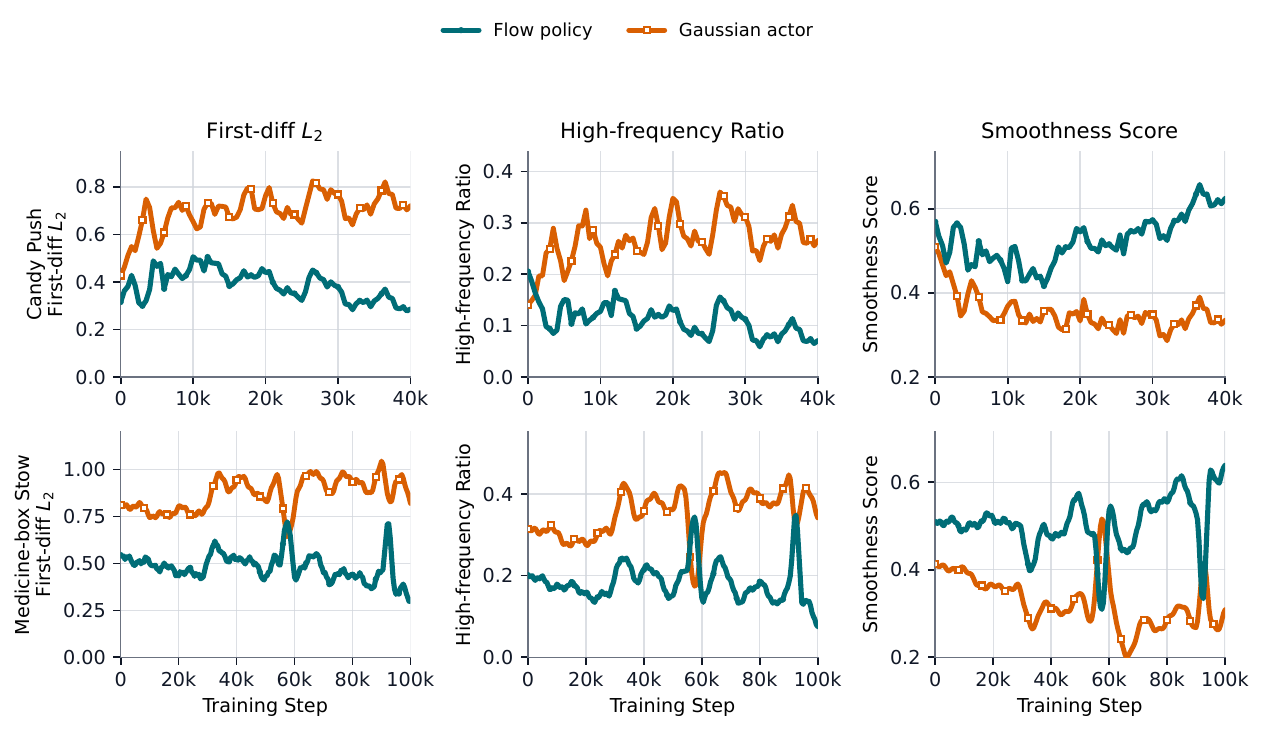}
\caption{Normalized-command smoothness over training for the flow policy
and a Gaussian actor on candy push and medicine-box stow. $D_1$ is the mean
consecutive-action $L_2$ change, $D_2$ is the mean second difference,
$P_{\mathrm{HF}}$ is the power fraction above half the Nyquist frequency, and
$S_{\mathrm{smooth}}$ is the aggregate score defined in
\eqref{eq:smooth-score}. Lower values of $D_1$, $D_2$, and $P_{\mathrm{HF}}$, and a
higher value of $S_{\mathrm{smooth}}$, indicate greater temporal regularity. These
single-run metrics do not measure physical jerk, safety, or human-perceived
predictability.}
\label{fig:afs-smooth}
\end{figure}

\FloatBarrier

\begin{table}[!t]
\centering
\caption{\IEEEtablecaptionfont AFS Objective Ablation on Candy Push under Original Lighting; SR: Success
Rate; T: Completion Time; $30$-Rollout Selected-Policy Results from One Run}
\label{tab:afs-component-ablation}
\footnotesize
\setlength{\tabcolsep}{3.8pt}
\begin{tabular}{@{}lccccc@{}}
\toprule
Method & FPO & BC & Smooth & SR $\uparrow$ & T $\downarrow$ \\
\midrule
No \AFS{}         & --         & --         & --         & 0.87 & 6.57 \\
\AFS{} w/o FPO    & --         & \checkmark & \checkmark & 0.93 & 7.24 \\
\AFS{} w/o BC     & \checkmark & --         & \checkmark & 0.87 & \textbf{\ensuremath{\mathbf{5.23}}} \\
\AFS{} w/o Smooth & \checkmark & \checkmark & --         & \textbf{\ensuremath{\mathbf{0.97}}} & 6.02 \\
Full \AFS{}       & \checkmark & \checkmark & \checkmark & \textbf{\ensuremath{\mathbf{0.97}}} & 5.44 \\
\bottomrule
\end{tabular}
\end{table}

\FloatBarrier

\paragraph{Reduced Command Irregularity}
Fig.~\ref{fig:afs-smooth} tracks three action-smoothness measures over training and
shows lower command irregularity for the flow policy in both tasks. For candy push,
the flow policy reduces the mean first-difference $L_2$ by $44.5\%$ ($0.387$
compared with $0.697$) and the high-frequency power ratio by $58.3\%$ ($0.112$
compared with $0.268$), while increasing the aggregate smoothness score by $49.0\%$.
For medicine-box stow, $D_1$ and $P_{\mathrm{HF}}$ decrease by $44.5\%$ ($0.483$
compared with $0.870$) and $47.9\%$ ($0.184$ compared with $0.353$), respectively,
while the aggregate score increases by $55.0\%$. The lower high-frequency content
is consistent with integrating the velocity field into a coherent action chunk and
applying the temporal-smoothness
term in \eqref{eq:afs-actor}. For the push task, the gap between the smoothness
curves also widens during training; this trend is descriptive because the curves do
not represent independent training runs.

\begin{figure}[!t]
\centering
\includegraphics[width=0.94\columnwidth]{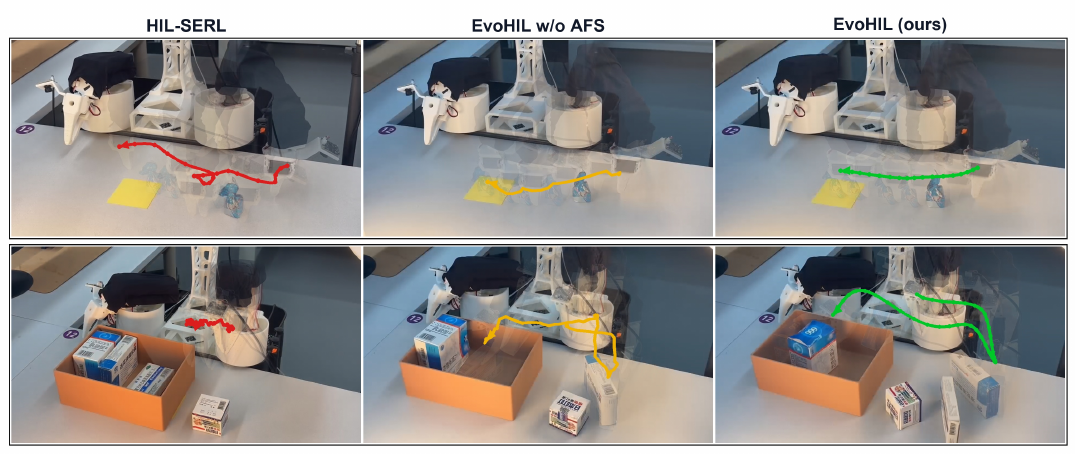}
\caption{Illustrative execution trajectories at the $60\%$ illumination shift on
the SO-101. Each panel overlays one end-effector path; arrows indicate execution
direction. These examples are not used as aggregate evidence.}
\label{fig:qualitative-trajectory}
\end{figure}

\paragraph{Illustrative Workspace Trajectories}
In Fig.~\ref{fig:qualitative-trajectory}, the complete system
follows a more direct path than HIL-SERL and the variant without \AFS{}. This visual
comparison is consistent with, but does not replace, the aggregate command metrics
in Fig.~\ref{fig:afs-smooth}.

\begin{figure}[!htbp]
\centering
\includegraphics[width=\columnwidth]{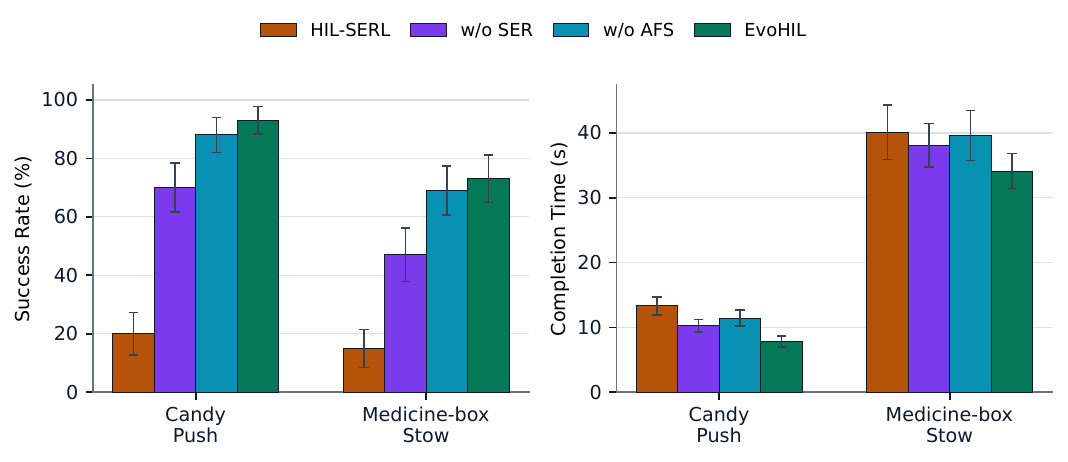}
\caption{Success rate and completion time for HIL-SERL, \EvoHIL{} without \SER{},
\EvoHIL{} without \AFS{}, and the complete \EvoHIL{} system on candy push and
medicine-box stow (SO-101) at the $60\%$ illumination shift. Error bars show the
binomial standard error over $30$ rollouts for success and the archived duration
spread for completion time. The run-level duration samples and the precise spread
statistic were not retained.}
\label{fig:afs-success}
\end{figure}

\paragraph{Contribution of the \AFS{} Components}
Table~\ref{tab:afs-component-ablation} evaluates the three \AFS{} objectives under
the original lighting. The variant without FPO records an SR of $0.93$ rather than
$0.97$; this difference is within the resolution of a $30$-rollout evaluation,
while its mean duration increases from $5.44$ to $7.24\,\mathrm{s}$. The variant
without BC has the shortest duration but an SR of $0.87$. The variant without
smoothness retains an SR of $0.97$ and records a duration $0.58\,\mathrm{s}$ longer
than the complete objective. These selected-policy results identify the best
observed success--duration balance; they do not isolate training-run effects.

\begin{figure*}[!t]
\centering
\includegraphics[width=\textwidth]{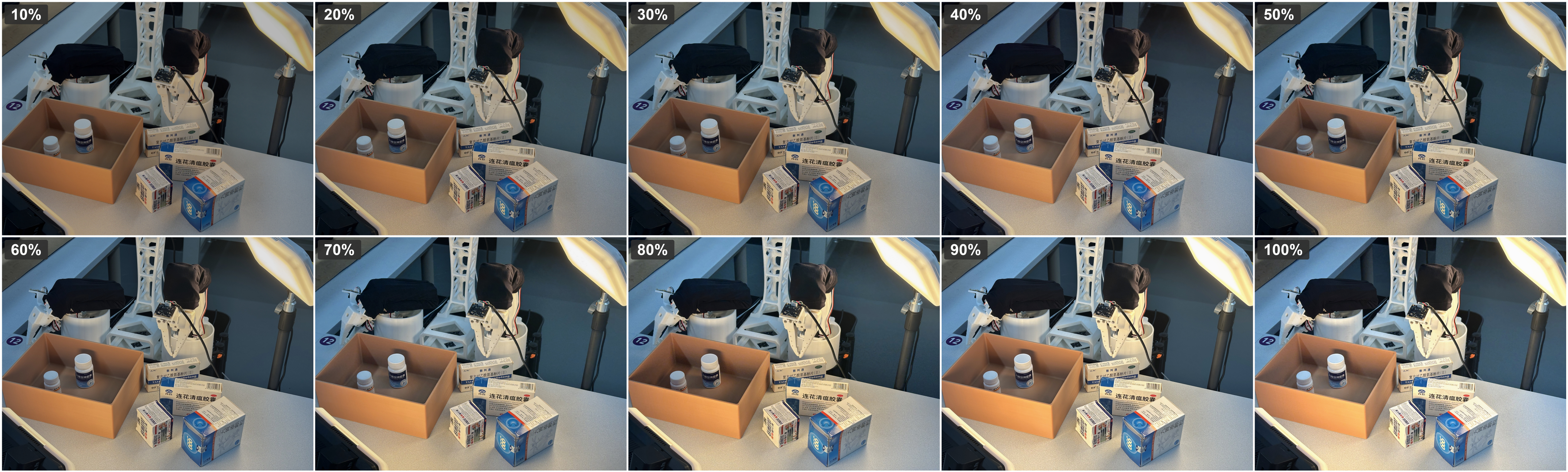}
\caption{The ten ordered illumination conditions used for evaluation, spanning
combinations of brightness, color temperature, shadow, and reflection changes. The
percentage labels form a composite test index rather than calibrated lux changes.}
\label{fig:lighting-grid}
\end{figure*}

\paragraph{Each Evolving Source Has a Distinct Role}
Fig.~\ref{fig:afs-success} compares the \SER{} and \AFS{} ablations with HIL-SERL.
Both
ablated variants improve over HIL-SERL, and the complete system performs best.
The observed differences suggest distinct roles. The larger success-rate change is
associated with \SER{}:
removing \AFS{} (keeping \SER{}) reaches $0.88$ on push and $0.69$ on stow, whereas
removing \SER{} (keeping \AFS{}) reaches $0.70$ and $0.47$. The larger duration
change is associated with \AFS{}: including \AFS{} reduces the
completion time from $11.45$ to $7.80\,\mathrm{s}$ for push, a reduction of $32\%$,
and from $39.60$ to $34.12\,\mathrm{s}$ for stow. This pattern agrees with the
normalized-command trends of Fig.~\ref{fig:afs-smooth}. The complete system shows
the largest improvement among the evaluated configurations.

\subsection{Multi-Embodiment Evaluation Under Illumination Shift}
\label{sec:exp-crossembodiment}
\begin{figure*}[!t]
\centering
\includegraphics[width=0.97\textwidth]{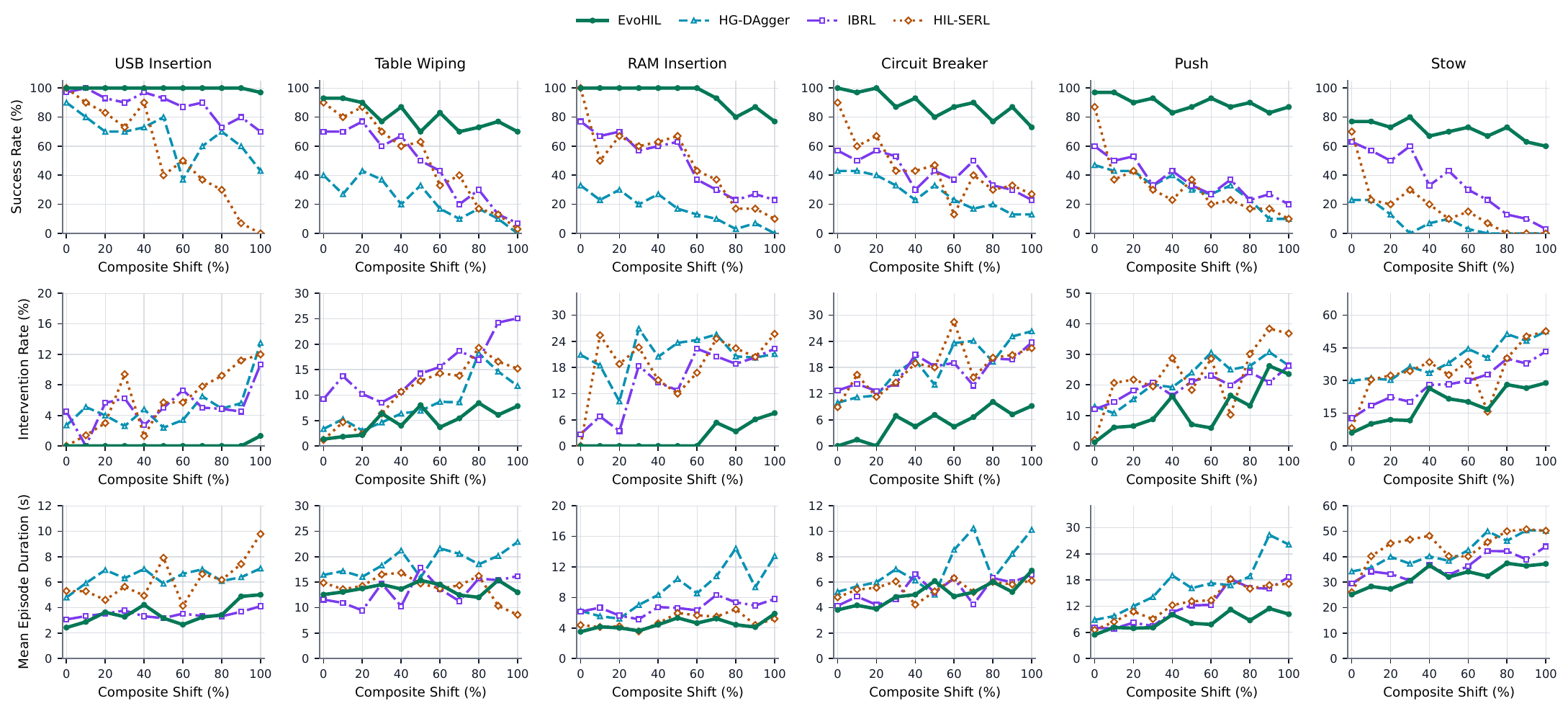}
\caption{Composite illumination-shift sweep ($0\%$ to $100\%$): success rate,
intervention rate, and mean recorded episode duration for \EvoHIL{} and the
baselines across six scenes. All \EvoHIL{} results use the unified SER--AFS
pipeline and its required retention-aware offline phase. Each point summarizes
$30$ rollouts of one selected policy and does not show training-run uncertainty.
BC and ACT are omitted because complete condition-wise sweep logs were not retained;
their $60\%$ results are reported in Table~\ref{tab:60shift}.}
\label{fig:lighting-gradient}
\end{figure*}

We next compare the complete \EvoHIL{} pipeline with all baselines under a
predefined illumination sweep. Fig.~\ref{fig:lighting-grid} shows the evaluation
conditions, and Fig.~\ref{fig:lighting-gradient} reports results across the complete
$0\%$--$100\%$ gradient for all six scenes. All \EvoHIL{} results use \SER{}, the
AFS action-chunk actor and execution-prefix critic, and the
retention-aware offline phase described in Section~\ref{sec:method-substrate};
task-specific horizons are listed in Table~\ref{tab:task-configs}.

\paragraph{Trends Across the Full Sweep}
The full sweep in Fig.~\ref{fig:lighting-gradient} places the result at the $60\%$
operating point in the context of the complete gradient. \EvoHIL{}
records high success rates across the range, with only gradual declines under the
most extreme shifts, whereas the baselines degrade sharply outside the training
condition. For USB insertion, \EvoHIL{} maintains a success rate of $1.00$ through
the $90\%$ shift and achieves $0.97$ at $100\%$. Over the same range, HIL-SERL
decreases from $1.00$ to $0.00$, and IBRL decreases from $0.97$ to $0.70$. For
candy push, HIL-SERL and \EvoHIL{} both achieve $0.97$ at $0\%$, but HIL-SERL
decreases to $0.37$ at a $10\%$ shift and to $0.10$ at $100\%$. \EvoHIL{} remains
above $0.83$ throughout the sweep. For medicine-box stow, HIL-SERL reaches $0.00$
from $80\%$ onward, whereas \EvoHIL{} remains between $0.60$ and $0.80$. The
intervention-rate and completion-time panels exhibit the same pattern. For example,
the intervention rate of \EvoHIL{} for USB insertion remains $0.00\%$ before
increasing to $1.32\%$ at $100\%$, compared with $12\%$ for HIL-SERL. The recorded
episode duration also remains nearly constant for \EvoHIL{} but increases for the
baselines. The same descriptive pattern therefore appears in intervention rate and
duration as well as success rate.

\begin{table*}[!t]
\centering
\caption{\IEEEtablecaptionfont Success Rate $\uparrow$ / Mean Episode Duration $\downarrow$ (s) at The
$60\%$ Illumination Shift; Each Cell Is A $60$-Trial Mean for One Selected Policy;
Row-Wise Best Values Are Bold; Duration Is Interpreted with Success}
\label{tab:60shift}
\footnotesize\setlength{\tabcolsep}{4pt}
\begin{tabular*}{\textwidth}{@{\extracolsep{\fill}}llcccccc@{}}
\toprule
Embodiment & Scene & HIL-SERL & HG-DAgger & BC & IBRL & ACT & \EvoHIL{} \\
\midrule
\multirow{4}{*}{FR3}
 & USB insertion     & 0.50 / 4.14 & 0.37 / 6.66 & 0.27 / 7.11 & 0.87 / 3.53 & 0.38 / 3.52 & \textbf{\ensuremath{\mathbf{1.00}}} / \textbf{\ensuremath{\mathbf{2.48}}} \\
 & RAM insertion     & 0.43 / 5.66 & 0.13 / 8.54 & 0.07 / \textbf{\ensuremath{\mathbf{4.99}}} & 0.37 / 6.31 & 0.30 / 6.23 & \textbf{\ensuremath{\mathbf{1.00}}} / 5.66 \\
 & Table wiping      & 0.33 / 13.74 & 0.17 / 21.65 & 0.13 / \textbf{\ensuremath{\mathbf{11.31}}} & 0.43 / 13.66 & 0.23 / 14.75 & \textbf{\ensuremath{\mathbf{0.87}}} / 14.39 \\
 & Circuit breaker   & 0.13 / 6.32 & 0.23 / 8.54 & 0.33 / 6.64 & 0.37 / 6.30 & 0.40 / 14.19 & \textbf{\ensuremath{\mathbf{0.83}}} / \textbf{\ensuremath{\mathbf{5.67}}} \\
\midrule
\multirow{2}{*}{SO-101}
 & Candy push        & 0.20 / 13.32 & 0.27 / 17.26 & 0.07 / 11.46 & 0.27 / 12.30 & 0.32 / 13.27 & \textbf{\ensuremath{\mathbf{0.93}}} / \textbf{\ensuremath{\mathbf{7.80}}} \\
 & Medicine-box stow & 0.15 / 40.12 & 0.03 / 42.59 & 0.02 / 41.02 & 0.30 / 36.34 & 0.45 / 36.01 & \textbf{\ensuremath{\mathbf{0.73}}} / \textbf{\ensuremath{\mathbf{34.12}}} \\
\bottomrule
\end{tabular*}
\end{table*}

\begin{table*}[!t]
\centering
\scriptsize
\setlength{\tabcolsep}{3.5pt}
\caption{\IEEEtablecaptionfont Appendix A: per-Task Setup and Hyperparameters; Demos, $T_{\max}$, and
$H/E$ Denote Initial Demonstrations, Maximum Episode Length, and
Chunk/Execution Horizons}
\label{tab:task-configs}
\begin{tabular*}{\textwidth}{@{\extracolsep{\fill}}llllccc@{}}
\toprule
Task & Embodiment & Action space & Cameras & Demos & $T_{\max}$ & $H/E$ \\
\midrule
USB insertion     & FR3    & 6-D EE delta twist; fixed gripper & wrist\_1, wrist\_2       & 20 & 150  & 8/2 \\
RAM insertion     & FR3    & 6-D EE delta twist; fixed gripper & wrist\_1, wrist\_2       & 20 & 150  & 8/2 \\
Table wiping      & FR3    & 6-D EE delta twist; fixed gripper & side, wrist\_1, wrist\_2 & 30 & 250  & 16/4 \\
Circuit breaker   & FR3    & 6-D EE delta twist; fixed gripper & wrist\_1, wrist\_2       & 20 & 150  & 8/2 \\
Candy push         & SO-101 & 6-D flow; xyz active, fixed gripper & top, side                & 30 & 600  & 24/4 \\
Medicine-box stow  & SO-101 & 7-D flow; xyz and thresholded gripper active & top, side, wrist & 30 & 2000 & 24/4 \\
\midrule
\multicolumn{7}{@{}p{0.98\textwidth}@{}}{Shared: \AFS{} with RLPD replay and a
ResNet encoder~\cite{he2016resnet}; equal replay/prior sampling; $\alpha{=}0.75$,
$\lambda_Q{=}0.2$, and $\lambda_\pi{=}0.1$.} \\
\multicolumn{7}{@{}p{0.98\textwidth}@{}}{Optimization: batch size $256$;
replay capacity $200$k; $\gamma{=}0.97$; Adam actor/critic learning rates
$10^{-4}/3{\times}10^{-4}$; gradient-norm limits $1/10$; target-update rate
$0.005$; two critics; pretrained ResNet-10 image encoder.} \\
\multicolumn{7}{@{}p{0.98\textwidth}@{}}{\SER{}: threshold $0.95$ for two frames;
negative-to-positive ratio $5$; baseline weight $0.1$; validation threshold $0.92$;
held-out fraction $0.25$; EMA rate $0.1$.} \\
\multicolumn{7}{@{}p{0.98\textwidth}@{}}{\AFS{}: five Euler steps; four flow-noise and
four value samples; $\varepsilon{=}0.1$; position decay $0.95$; imitation weight
$0.2$ with decay factor $0.5$; smoothness weight $0.05$.} \\
\multicolumn{7}{@{}p{0.98\textwidth}@{}}{FPO: $5$k-update warm-up and $5$k-update
ramp; disabled during retention-aware offline fine-tuning.} \\
\bottomrule
\end{tabular*}
\vspace{-4pt}
\end{table*}

\begin{table}[!t]
\centering
\scriptsize
\setlength{\tabcolsep}{2.6pt}
\caption{\IEEEtablecaptionfont Appendix B: Relighting Cost for One $8{,}000$-Transition Camera Stream
on One GPU (BF16); The Lower-Cost Backend Is Used}
\label{tab:relight-cost}
\begin{tabular*}{\columnwidth}{@{\extracolsep{\fill}}lccc@{}}
\toprule
Backend & Time (h) $\downarrow$ & Memory (GB) $\downarrow$ & Consistency \\
\midrule
Cosmos-Transfer1~\cite{nvidia2025cosmostransfer} & 6.82 & 26.80 & Reference \\
Uni-Relight~\cite{he2025unirelight} & 48.63 & 39.88 & Higher \\
\bottomrule
\end{tabular*}
\end{table}

\begingroup
\setlength{\parskip}{0pt}
\paragraph{Comparison at the 60\% Operating Point}
Table~\ref{tab:60shift} reports the $60\%$ shift. \EvoHIL{} records the highest
success rate for every scene and embodiment. On the FR3, the corresponding rates
are $1.00$, $1.00$, $0.87$, and $0.83$ for USB insertion, RAM insertion, table
wiping, and circuit breaker, compared with $0.50$, $0.43$, $0.33$, and $0.13$ for
HIL-SERL. IBRL reaches $0.87$ on USB insertion, while the mean durations are
$3.53\,\mathrm{s}$ for IBRL and $2.48\,\mathrm{s}$ for \EvoHIL{}. On the SO-101,
the rates of \EvoHIL{} are $0.93$ for candy push and $0.73$ for medicine-box stow,
compared with $0.32$ and $0.45$ for ACT. The corresponding mean durations are
$7.80$ and $34.12\,\mathrm{s}$ for \EvoHIL{}, and $13.27$ and
$36.01\,\mathrm{s}$ for ACT.

Success and duration do not improve uniformly. BC records the shortest durations
for RAM insertion and table wiping, but its success rates are only 0.07 and 0.13,
compared with 1.00 and 0.87 for \EvoHIL{}. On the remaining four scenes,
\EvoHIL{} achieves both the highest success and the shortest duration. A short
episode can result from early failure rather than efficient task completion;
duration must therefore be interpreted together with success. This distinction
prevents short failed trials from being interpreted as faster task execution.
\par
\endgroup

\section{Conclusion and Limitations}
\label{sec:conclusion}
\EvoHIL{} adapts the reward, action, and visual interfaces within a human-in-the-loop
actor--critic workflow. \SER{} updates the success classifier under label-source
isolation, \AFS{} learns action chunks with an execution-prefix critic, and the
retention-aware offline phase uses relit interaction records with frozen AFS
references. The components interact through the critic, leaving the sparse reward
semantics unchanged.
Across six tasks, the selected policies yield higher descriptive success rates
than the baselines. Proxy-agreement and normalized-command results support the
intended component roles.

Several limitations remain. \SER{} measures agreement with a confirmation proxy
rather than latent success. The empirical CFM-loss ratio of FPO is not a PPO
importance ratio. Relit replay assumes preserved geometry, state, and temporal
correspondence. The sweep targets adaptation rather than zero-shot generalization;
only \EvoHIL{} uses relit observations and offline updates, so the comparison is not
resource-matched.

\FloatBarrier
\appendices
The adopted relighting backend is external to \EvoHIL{}; its fidelity limits the
illumination coverage of replay. The same transition-preserving interface can
accommodate another relighter without changing \SER{}, \AFS{}, replay mixing, or the
anchors.

\FloatBarrier

\FloatBarrier
\balance
\bibliographystyle{IEEEtran}
\bibliography{references}

\end{document}